\def\year{2021}\relax
\documentclass[letterpaper]{article} 
\usepackage{aaai21}  
\usepackage{times}  
\usepackage{helvet} 
\usepackage{courier}  
\usepackage[hyphens]{url}  
\usepackage{graphicx} 
\def\UrlFont{\rm}  
\usepackage{natbib}  
\usepackage{caption} 
\usepackage{enumitem}
\usepackage{multirow}
\usepackage{booktabs}
\usepackage{subcaption}
\usepackage{xr}
\usepackage{bbold}
\usepackage{booktabs}
\usepackage{amsmath}
\usepackage{xcolor,pifont}
\usepackage{MnSymbol,wasysym}
\usepackage{fontawesome}
\usepackage{graphicx}
\usepackage{anyfontsize}
\usepackage{tikz}
\usepackage{todonotes}
\usepackage{array}
\newcommand{\telin}[1]{{\color{blue}{\small\bf\sf [Te-Lin: #1]}}}
\newcommand{\nanyun}[1]{{\color{red}{\small\bf\sf [Nanyun: #1]}}}

\newcommand{\xiangci}[1]{{\color{cyan}{\small\bf\sf [Xiangci: #1]}}}
\newcommand{\nuan}[1]{{\color{cyan}{\small\bf\sf [Nuan: #1]}}}

\newcommand{\Skip}[1]{}

\newcommand{\ie}{\textit{i}.\textit{e}.\ }
\newcommand{\eg}{\textit{e}.\textit{g}.\ }
\newcommand{\dataset}{\textsc{Melinda}}

\newcommand{\datasetsrc}{\textbf{M}ultimodal biom\textbf{E}dica\textbf{L} exper\textbf{I}me\textbf{N}t metho\textbf{D} cl\textbf{A}ssification}

\newcommand{\secref}[1]{Section \ref{#1}}
\newcommand{\figref}[1]{Figure \ref{#1}}
\newcommand{\tbref}[1]{Table \ref{#1}}

\newcommand{\dotieconcat}[2]{
  \text{\raisebox{.8ex}{$\smallfrown$}}%
}

\newcommand{\mypar}[1]{\noindent\textbf{#1}}

\title{\dataset: A Multimodal Dataset \\for Biomedical Experiment Method Classification}

\author {
    Te-Lin Wu\textsuperscript{\rm 1},
    Shikhar Singh\textsuperscript{\rm 2},
    Sayan Paul\textsuperscript{\rm 3},
    Gully Burns\textsuperscript{\rm 4},
    Nanyun Peng\textsuperscript{\rm 1} \\
}
\affiliations {
    \textsuperscript{\rm 1} University of California, Los Angeles,  \textsuperscript{\rm 2}  University of Southern California \\
    \textsuperscript{\rm 3} Intuit Inc., \textsuperscript{\rm 4} Chan Zuckerberg Initiative \\
    \{telinwu, violetpeng\}@cs.ucla.edu\textsuperscript{\rm 1}, ssingh43@usc.edu\textsuperscript{\rm 2},
    sayan.paul6@gmail.com\textsuperscript{\rm 3}, gully.burns@chanzuckerberg.com\textsuperscript{\rm 4}
}

\begin{document}

\maketitle

\begin{abstract}
    We introduce a new dataset, \dataset, for \datasetsrc. 
The dataset is collected in a fully automated \textit{distant supervision} manner, where the labels are obtained from an existing curated database, and the actual contents are extracted from papers associated with each of the records in the database.
We benchmark various state-of-the-art NLP and computer vision models, including unimodal models which only take either caption texts or images as inputs, and multimodal models. Extensive experiments and analysis show that multimodal models, despite outperforming unimodal ones, still need improvements especially on a less-supervised way of grounding visual concepts with languages, and better transferability to low resource domains. 
We release our dataset and the benchmarks to facilitate future research in multimodal learning, especially to motivate targeted improvements for applications in scientific domains.

\end{abstract}

\section{Introduction}
\label{sec:intro}

Biocuration, the activity of \textit{manually} organizing biological information, is a crucial yet human-effort-intensive process in biomedical research~\cite{10.1371/journal.pbio.2002846}. 
Organizing such knowledge in a structured way is important for accelerating science since it facilitates downstream tasks such as scientific information retrieval~\cite{craven1999constructing, mohan2018fast, burns2018towards, burns2019building}, and question answering~\cite{ImageCLEFVQA-Med2019, nguyen2019overcoming,he2020pathvqa}. 

\begin{figure}[t]
\centering
    \includegraphics[width=1\columnwidth]{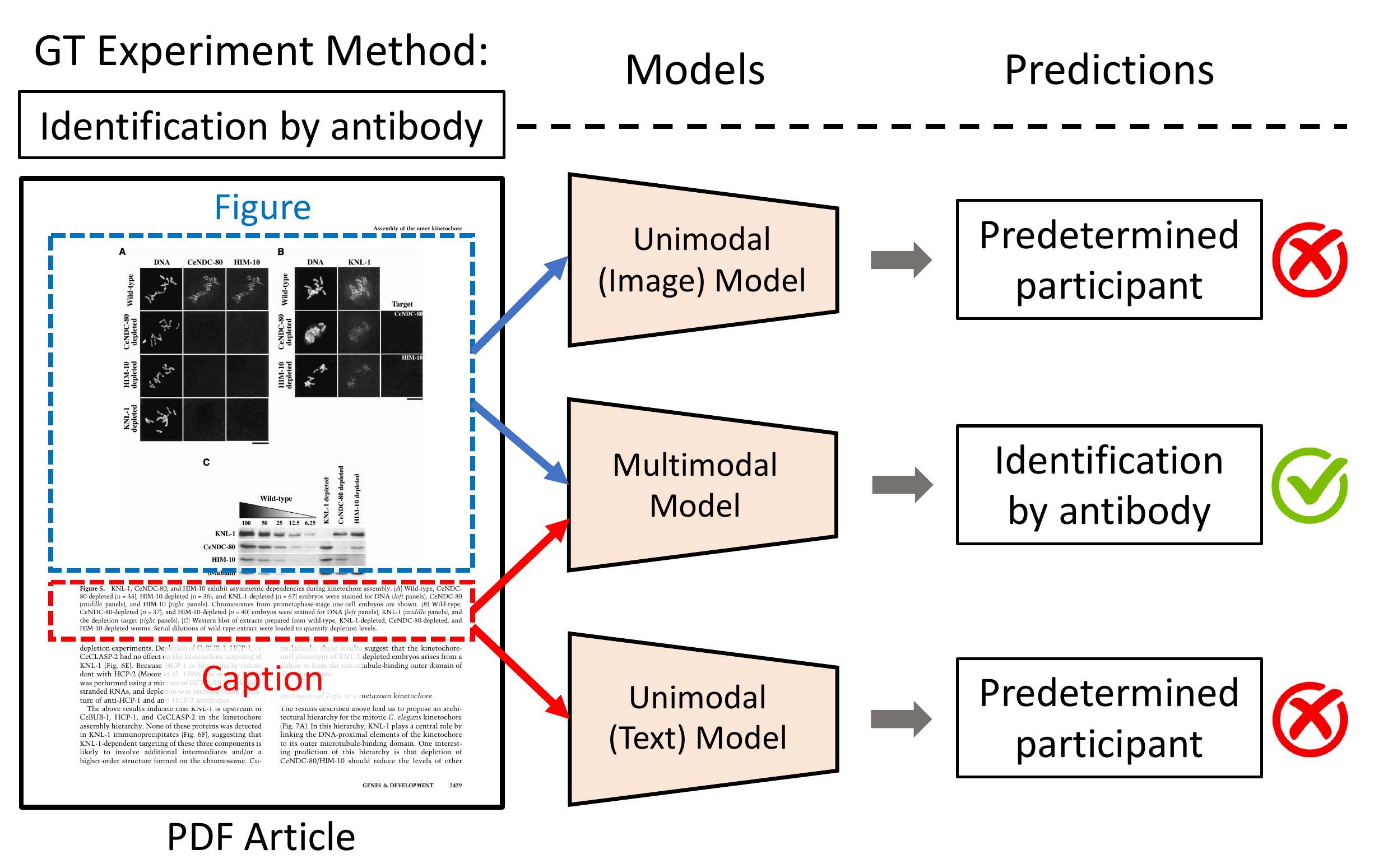}
    \caption{
        \textbf{The \dataset~dataset \& the biomedical experiment classification task:}
        We introduce a new dataset which concerns learning to recognize the underlying experiment methods used to produce an experimental figure in biomedical research articles. The recognition is fundamentally multimodal, where justification of the experiment methods takes both figures and captions into consideration. The \dataset~dataset could serve as a good testbed for benchmarking, as well as motivating multimodal models particularly in biomedical and low-resource domains.
    }
    \label{fig:teaser}
\end{figure}

One such curation task is recognizing \textit{experiment methods}, which identifies the underlying experimental protocols that result in the figures in research articles. 
It can be formulated as a multi-class classification task, which takes as inputs the figures and their captions, and outputs the corresponding experiment types that generate the figures, as illustrated in~\figref{fig:teaser}.

The task is inherently \textit{multimodal} as biocurators need to take both the figure and the caption into consideration to make their decisions~\cite{demner2012design}.\footnote{Although different experiment methods tend to generate visually different results, the differences can be subtle and the captions often help distinguish these subtle differences among figures.}
While scientists can do the task with perfect accuracy, the requirements of manual labeling from experts hinder the scalability of the process. It is thus imperative to develop advanced language and computer vision multimodal tools to help 
accelerate the aforementioned scientific discovery process.

However, automatically identifying the experiment methods poses significant challenges for multimodal processing tools. One major challenge is how to ground the visual concepts to language. Most current visual-linguistics multimodal models~\cite{li2019visualbert,lu2019vilbert,su2019vl,chen2019uniter} rely on a robust \textit{object detection} module to identify \textit{predefined} objects for grounding finer granularity of visual and linguistics concepts. However, as it requires extra efforts from experts, scientific images often lack ground truth object annotations, and the transfer of pretrained detection models suffers from significant domain shifts. As a result, this specific domain would appreciate multimodal models particularly with less-supervised grounding paradigms.
In addition, it is expensive to collect annotations from domain experts; the lack of sizable benchmark datasets hinders the development of multimodal models tailored to the biomedical domain.

To spur research in this area, we introduce \dataset{}, a dataset for \datasetsrc{} that is created through a fully automated \textit{distantly supervised} process~\cite{mintz2009distant}. Specifically, we leverage an existing biomedical database, \textbf{IntAct}\footnote{https://www.ebi.ac.uk/intact/}~\cite{orchard2013mintact}, to get the experiment method labels, and then properly extract the actual contents from papers pointed by the records in IntAct to pair with the obtained labels.
\dataset{} features 2,833 figures paired with their corresponding captions.
We further segment captions into sub-captions referring to different sub-figures in the images, resulting in a total of 5,371 data records along with the labels of the experiment methods used to generate the sub-figures.

We benchmark several state-of-the-art models on the proposed experiment method classification task, including unimodal vision and language models and multimodal ones. 
Experiments suggest that multimodality is helpful for achieving better performances.
However, the performances are still far from expert human-level, which suggests several area of improvements, including less reliance on object detection for grounding linguistic representations with visual sources, as well as finer-grained multimodal groundings.

Our work sheds light on future research in: (1) more generally applicable multimodal models, and (2) better transfer learning techniques in low resource domains such as scientific articles~\cite{gururangan2020don}. 
We summarize our main contributions as follows:
\begin{itemize}[leftmargin=*]
    \item A multimodal dataset mapping compound figures and associated captions from biomedical research articles to the labels of experiment methodologies
    , to help spur the research on multimodal understanding for scientific articles. 
    \item We conducted extensive experiments to benchmark and analyze various unimodal and multimodal models against the proposed dataset, suggesting several future directions for multimodal models in scientific domain. 
\end{itemize}

\section{The \dataset~Dataset} 
\label{sec:dataset}





We introduce a new multimodal dataset, \dataset, for biomedical experiment method classification.
Each data instance is a unique tuple consisting of a figure, an associated sub-caption for the targeted sub-figure(s), and an experiment method label coming from the IntAct database.
IntAct stores manually annotated labels for experiment method types, paired with their corresponding sub-figure identifiers and ids to the original paper featuring the figures\footnote{IntAct only stores these ids as pointers, and our collection pipeline extracts the actual contents according to these pointers.}, and structures them into an ontology. Each major category have different levels of granularity. This work mainly focuses on two major categories of experiments for identifying molecular interactions: \textit{participant identification (Par)} and \textit{interaction detection (Int)} methods\footnote{Molecular interaction experiments require two types of assay: participant detection methods identify the molecules involved in the interaction and the interaction detection methods identify the types of interactions occurring between the two molecules.
}, each has two levels of granularity, coarse and fine (choice of the granularity depends on downstream applications).
Samples of data and their labels are as exemplified in~\figref{fig:data_samples} (more are in the appendix).

Each record in IntAct consists of the aforementioned expert curated information to a specific article in the \textbf{Open Access PubMed Central}\footnote{A publicly available subset of the PubMed collections: https://www.ncbi.nlm.nih.gov/pmc/tools/openftlist/} (\textbf{OA-PMC}).
According to the IntAct guideline, figure captions are sufficiently descriptive for justifying the underlying methods of the figures, and hence are properly extracted instead of including the body of text in the articles.
The details of the dataset collection procedures and its statistics are described in the following sections.

\begin{figure}[t!]
\begin{subtable}{.47\columnwidth}

{\tiny \textbf{\texttt{Experiment Method Labels}}}

\vspace{1mm}

\fontsize{2mm}{5} \selectfont \texttt{Par(coarse)\hspace{1mm}:\hspace{0.5mm}predetermined}

\fontsize{2mm}{5} \selectfont \texttt{\hspace{15mm} participant}

\vspace{1mm}
\fontsize{2mm}{5} \selectfont \texttt{Par(fine)\hspace{3.5mm}:\hspace{0.5mm}predetermined}

\fontsize{2mm}{5} \selectfont \texttt{\hspace{15mm} \space}

\vspace{1mm}
\fontsize{2mm}{5} \selectfont \texttt{Int(coarse)\hspace{1mm}:\hspace{0.5mm}imaging}

\fontsize{2mm}{5} \selectfont \texttt{\hspace{15mm} technique}

\vspace{1mm}
\fontsize{2mm}{5} \selectfont \texttt{Int(fine)\hspace{3.5mm}:\hspace{0.5mm}fluorescence}

\fontsize{2mm}{5} \selectfont \texttt{\hspace{15mm} imaging}

\vspace{2mm}

\centering
  \includegraphics[width=\columnwidth]{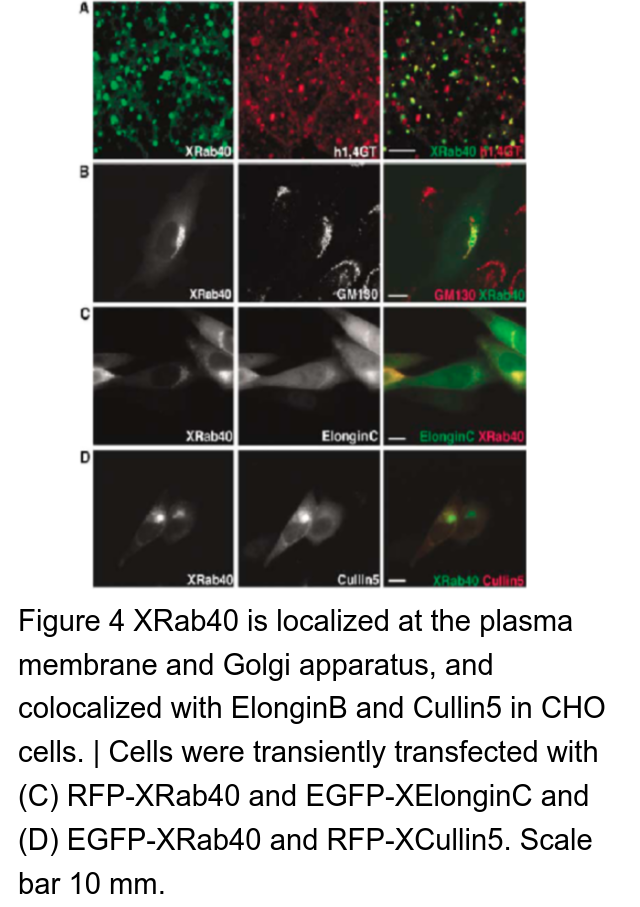}
\end{subtable}%
\quad
\begin{subtable}{.47\columnwidth}

{\tiny \textbf{\texttt{Experiment Method Labels}}}

\vspace{1mm}

\fontsize{2mm}{5} \selectfont \texttt{Par(coarse)\hspace{1mm}:\hspace{0.5mm}identification}

\fontsize{2mm}{5} \selectfont \texttt{\hspace{15mm} by antibody}

\vspace{1mm}

\fontsize{2mm}{5} \selectfont \texttt{Par(fine)\hspace{3.5mm}:\hspace{0.5mm}western}

\fontsize{2mm}{5} \selectfont \texttt{\hspace{15mm} blot}

\vspace{1mm}

\fontsize{2mm}{5} \selectfont \texttt{Int(coarse)\hspace{1mm}:\hspace{0.5mm}affinity}

\fontsize{2mm}{5} \selectfont \texttt{\hspace{14.5mm} chromatography}

\vspace{1mm}

\fontsize{2mm}{5} \selectfont \texttt{Int(fine)\hspace{3.5mm}:\hspace{0.5mm}anti-bait}

\fontsize{2mm}{5} \selectfont \texttt{\hspace{15mm} coip}

\vspace{2mm}

\centering
  \includegraphics[width=\columnwidth]{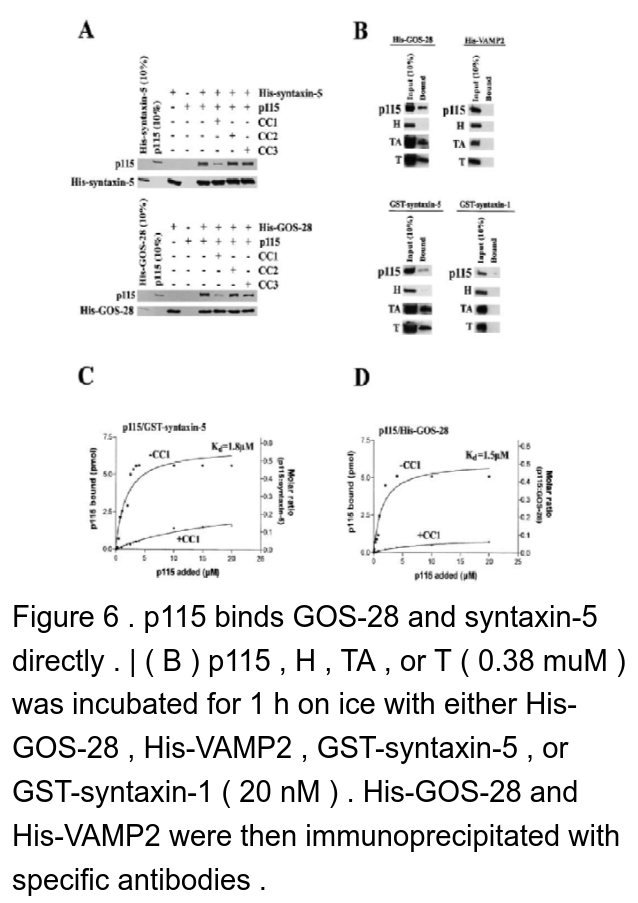}
 
\end{subtable}

\caption{
    \textbf{Sample data:} The basic structure of the data in \dataset~is composed of a figure, a sub-caption associated to one or multiple sub-figure(s), and a set of curated experiment method labels as shown on top of each figure. These labels represent the types of experiments conducted to generate the shown resulting sub-figures and captions.
    The above left sample concerns sub-figures (C) and (D), while the right sample concerns sub-figure (B), as indicated in their captions.
    Human experts tend to determine the labels leveraging features such as scientific terms concerning assays and methodologies in the captions, as well as indicative image features such as blots, graphs, and microscopic images.
}
\label{fig:data_samples}
\end{figure}















 


\subsection{Data Collection Pipeline}
\label{ssec:collection_pipeline}

\begin{figure*}[ht!]
\centering
    \includegraphics[width=0.87\textwidth]{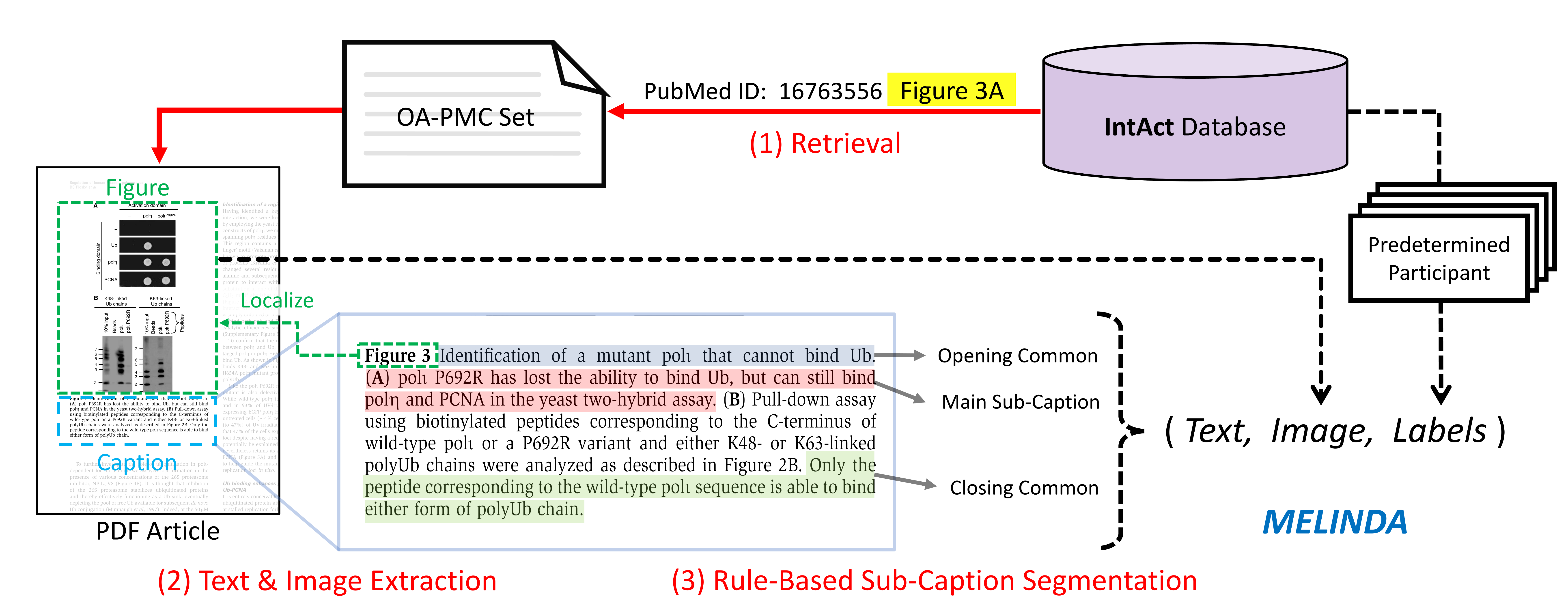}
    \caption{
        \textbf{Data collection pipeline}: Our collection pipeline is \textit{distantly supervised} and fully automatable. It consists of three main steps:
        \textbf{(1)} Retrieve the PDF article in the OA-PMC set using the PubMed id from the IntAct database.
        \textbf{(2)} Extract the caption blocks using an in-house PDF interpreter, and localize the nearby corresponding figures.
        \textbf{(3)} Segment the caption blocks into sub-captions.
        Combining all three steps with the paired labels gives a single data record in our \dataset~dataset.
    }
    \label{fig:pipeline}
\end{figure*}

Our dataset is collected through three main procedures, as illustrated in~\figref{fig:pipeline}:
(1) Obtain the experiment method labels and sub-figure identifiers from IntAct.
(2) Localize the indicated figures and their captions in the pointed PDF articles.
(3) Segment the captions into sub-captions so each can target a sub-figure of the figures obtained in step (2).
As the overall procedure adopts a fully automated \textit{distant supervision} approach, our dataset could be \textit{seamlessly expanded as additional articles being added} to the OA-PMC set.


\vspace{0.3em}
\mypar{Ground Truth IntAct Label Extraction.}
By properly parsing and mapping the {\em PSI-MI2.5}\footnote{An XML format: http://psidev.info/mif} formatted IntAct records, each individually extracted instance can form a unique tuple of $(\textit{experiment-method-label\textbf{s}}, \textit{sub-figure-id})$, where the \textit{sub-figure-id} is a concatenation of the PubMed id of an article and the sub-figure identifier.

\vspace{0.3em}
\mypar{Text and Image Extraction.}
The OA-PMC paper ids are then used to search and download the indicated PDF articles.
The textual and image contents are extracted using an in-house PDF interpreter,
which leverages spatial indexing over each page to support content extractions.
We extract contiguous word blocks across the articles, and the figure captions are localized by detecting the keywords `Fig' or `Figure'. 
The corresponding figures are cropped out by searching for large rectangular regions with low text densities nearby the captions.
Note that although the classification task concerns sub-figures, \textbf{we do not further segment a figure into sub-figures} as we expect the models to be equipped with the capability of attending to the right sub-figures given the captions. Moreover, there are captions cross-referencing multiple sub-figures, and thus full figures should be preserved.

\vspace{0.3em}
\mypar{Sub-Caption Segmentation.}
Captions for compound figures are first tokenized into sentences followed by a text cleansing preprocessing, and then grouped into proper corresponding sub-captions through the following steps:
(1) Descriptions before the first sentence containing sub-figure identifiers, \eg "(A)", "(A-C)", are extracted as the \textit{opening common} text.
(2) The sentence containing a detected sub-figure identifier and all of its subsequent ones until the next sentence containing different identifier(s) is found, are extracted as the \textit{main} sub-caption for that particular identifier. 
(3) Descriptions after the last sentence containing identifiers, are regarded as the \textit{closing common} text, as researchers may put some summary texts at the end.
Hence, a proper sub-caption is a concatenation of all of the above, which ensures no relevant contents of a sub-caption is overlooked.
More details of our data collection pipeline can be found in the appendix and our released code repository\footnote{The data collection pipeline and our benchmark models can be found at \url{https://github.com/PlusLabNLP/melinda}.}. 

\newcolumntype{L}{>{\arraybackslash}m{5cm}}

\begin{table}[t!]
\begin{subtable}{\columnwidth}
\centering
\small
    \begin{tabular}{cLc}
    \toprule
    \textbf{Quality} & \multicolumn{1}{c}{\textbf{Descriptions}} & \textbf{\%} \\
    \midrule
    \multirow{-1.2}{*}{\Large \faFrownO}
    & Imperfect crop of the figures, \ie accidentally cropped out some parts & 8 \\
    \midrule
    \multirow{3.5}{*}{\Large \faSmileO}
    & Perfect crop of figures but with some small additional nuisances \eg partial captions, other figures, etc. & 34 \\
    \cline{2-3} \\[-1em]
    & Perfect nuisance-free crop of figures with proper boundaries & 58 \\
    \bottomrule
    \end{tabular}
\caption{Image cropping quality assessments}
\label{tab:caption-qa1}
\end{subtable}

\begin{subtable}{\columnwidth}
\centering
\small
    \begin{tabular}{cLc}
    \\
    \toprule
    \textbf{Quality} & \multicolumn{1}{c}{\textbf{Descriptions}} & \textbf{\%} \\
    \midrule
    \multirow{-1.2}{*}{\Large \faFrownO}
    & Extracted captions do not match the original captions in the PDF \texttt{or} the extracted figures (caption-figure mismatch) & 4 \\
    \midrule
    \multirow{5}{*}{\Large \faSmileO}
    & Extracted and segmented sub-captions match the original sub-captions in the PDF \texttt{and} caption-figure matched & 10 \\
    \cline{2-3} \\[-1em]
    & Sub-captions matched the original sub-captions in the PDF with \textit{common} parts preserved \texttt{and} caption-figure matched & 86 \\
    \bottomrule
    \end{tabular}
\caption{Caption extraction \& segmentation quality assessments}
\label{tab:caption-qa2}
\end{subtable}
\caption{\textbf{Data quality assessments out of 100 random samples:} For both \textbf{(a)} image cropping and \textbf{(b)} caption extraction and segmentation, the assessments show there are over $90\%$ of samples regarded as good (\faSmileO), while there is a small proportion with certain noises in the extractions (\faFrownO).
}
\label{tab:caption-qa}
\end{table}

\subsection{Data Quality Assessment}
\label{ssec:dataset_qa}

Since our dataset is created by distant supervision from IntAct, for which if we perfectly pair the labels with corresponding figures and subcaptions, the \textit{expert} human performances should remain $\sim$100\%. Therefore, the quality of the data instances rely on the quality of content extraction and pairing. 
In order to estimate the quality of the extracted contents, we randomly sample 100 instances for a manual inspection.
With the corresponding original papers provided, we ask
three non-domain-expert annotators to assess the quality
mainly in terms of how good the image cropping is and how accurate the caption extractions are (the results were computed via majority vote). The inter-annotator agreement Fleiss' Kappa for the following results are 0.804 for images and 0.676 for captions assessments.

~\tbref{tab:caption-qa1} shows the inspection results of the extracted (and cropped) images on if they are missing any important regions, or containing any noises. Among the sampled images, $92\%$ (\ie $34+58$) of the images are showing reasonably good quality, with $8\%$ of them missing some details due to the cropping.
The quality of the extracted (and segmented) sub-captions, as well as whether they match the associated sub-figure images, is summarized in~\tbref{tab:caption-qa2}. Over $96\%$ of the sampled data can be regarded as good, while $4\%$ of them have issues such as partial texts missing. It is worth noting that even in this proportion of data which misses some details, the majority parts of the captions (and the figures) are still properly preserved.

\subsection{Dataset Details}
\label{ssec:data_details}

\paragraph{General Statistics.}

\begin{table}[t]
\centering
\small
\begin{tabular}{lrrrr}
    \toprule
    \multicolumn{1}{c}{\textbf{Type}} & \multicolumn{4}{c}{\textbf{Counts}} \\
    \midrule
    \multicolumn{1}{c}{Total Unique Articles} & \multicolumn{4}{c}{1,497} \\
    \multicolumn{1}{c}{Total Unique Images} & \multicolumn{4}{c}{2,833} \\
    \multicolumn{1}{c}{Total Data Instances} & \multicolumn{4}{c}{5,371} \\
    \multicolumn{1}{c}{Train / Val / Test} & \multicolumn{4}{c}{4,344 / 449 / 578} \\
    \multicolumn{1}{c}{Type-Token Ratio} & \multicolumn{4}{c}{29,384 / 501,091 = 0.059} \\
    \toprule
    \multicolumn{1}{c}{\textbf{Type}} & \multicolumn{1}{c}{\textbf{Mean}} & \multicolumn{1}{c}{\textbf{Std}} & \multicolumn{1}{c}{\textbf{Min}} & \multicolumn{1}{c}{\textbf{Max}} \\
    \midrule
    \multicolumn{1}{l}{Tokens in a Caption} & 93.29 & 47.33 & 3 & 491 \\
    \multicolumn{1}{l}{Sentences in a Caption} & 5.23 & 2.36 & 1 & 27  \\
    \multicolumn{1}{l}{Tokens in a Sentence}  & 17.83 & 12.27 & 1 & 256 \\
    \bottomrule
\end{tabular}
\caption{\textbf{General statistics of \dataset:} We provide the detailed component counts of our dataset, including the sizes for each split~\textbf{(upper half)}, and the statistics of tokens and sentences from the captions~\textbf{(lower half)}.}
\label{tab:data-det}
\end{table}

\begin{table}[t!]
\centering
\small
    \begin{tabular}{cccc}
    \toprule
    \begin{tabular}[c]{@{}l@{}}\textbf{Method}\\  \textbf{Category}\end{tabular} & \multicolumn{1}{c}{\textbf{Hierarchy}} & \multicolumn{1}{c}{\textbf{\# IntAct Labels}} &
    \begin{tabular}[c]{@{}l@{}}\textbf{\# Labels in} \\ \textbf{Our Dataset}\end{tabular} \\
    \midrule
    \multirow{2}{*}{Participant}
    & Coarse & 7 & 7 \\ \cline{2-4}
    \\ [-1em]
    & Fine & 48 & 45 \\
    [-.2em]
    \midrule
    \multirow{2}{*}{Interaction}
    & Coarse & 18 & 15 \\ \cline{2-4}
    \\ [-1em]
    & Fine & 122 & 85\\
    [-.2em]
    \bottomrule
    \end{tabular}
\caption{\textbf{Unique IntAct label counts:} For each of the main categories, \textit{participant} and \textit{interaction}, we list the number of unique labels in the original IntAct database and our collected dataset, for both the coarse and fine-grained labels.}
\label{tab:exp-type}
\end{table}

\begin{figure}[tb!]
\begin{subtable}{.47\columnwidth}
\centering
  \includegraphics[width=0.88\columnwidth]{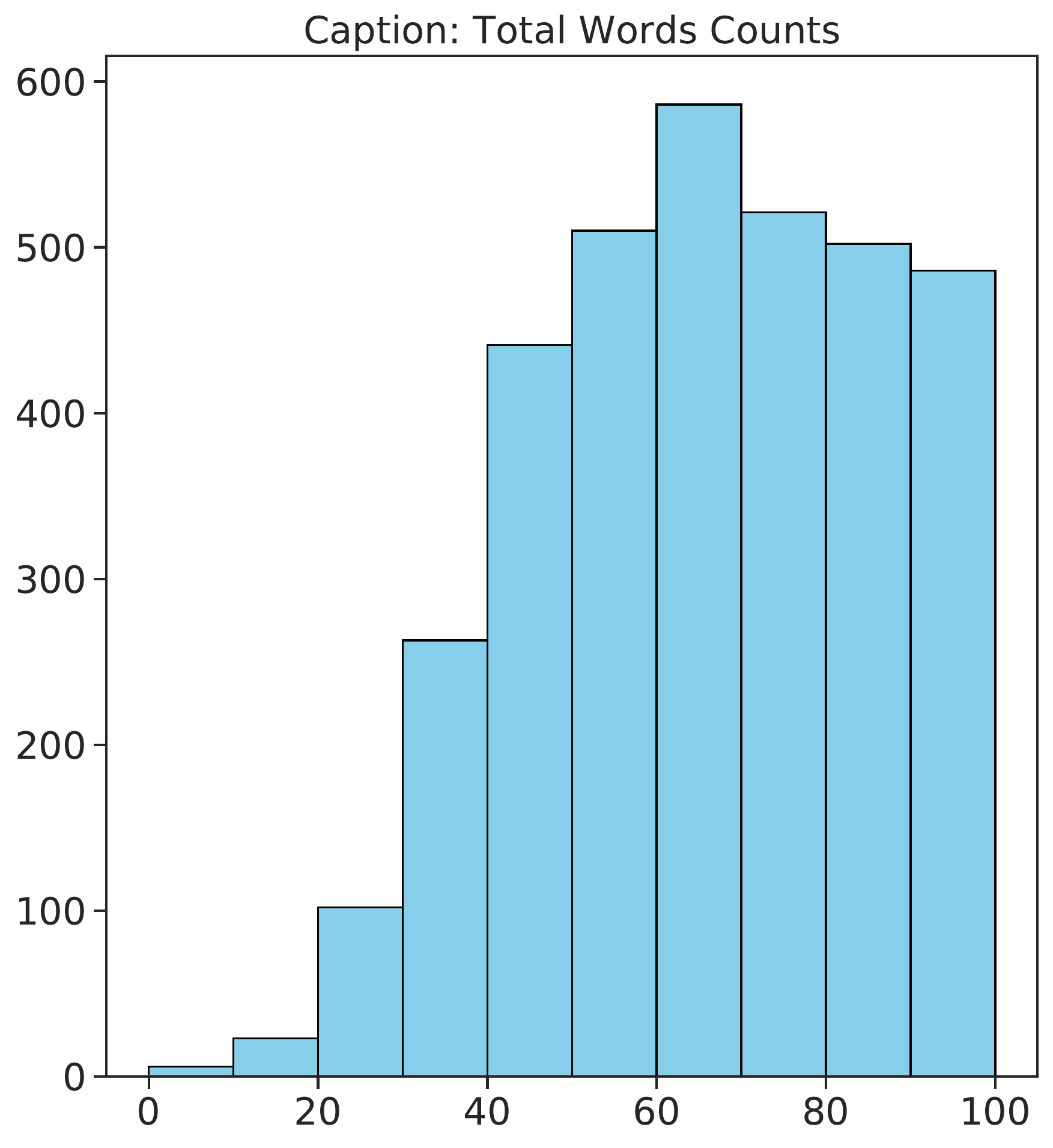}
    \caption{
        Word Counts.
    }
    \label{fig:data_wsc_1}
\end{subtable}%
\quad
\begin{subtable}{.47\columnwidth}
\centering
  \includegraphics[width=0.88\columnwidth]{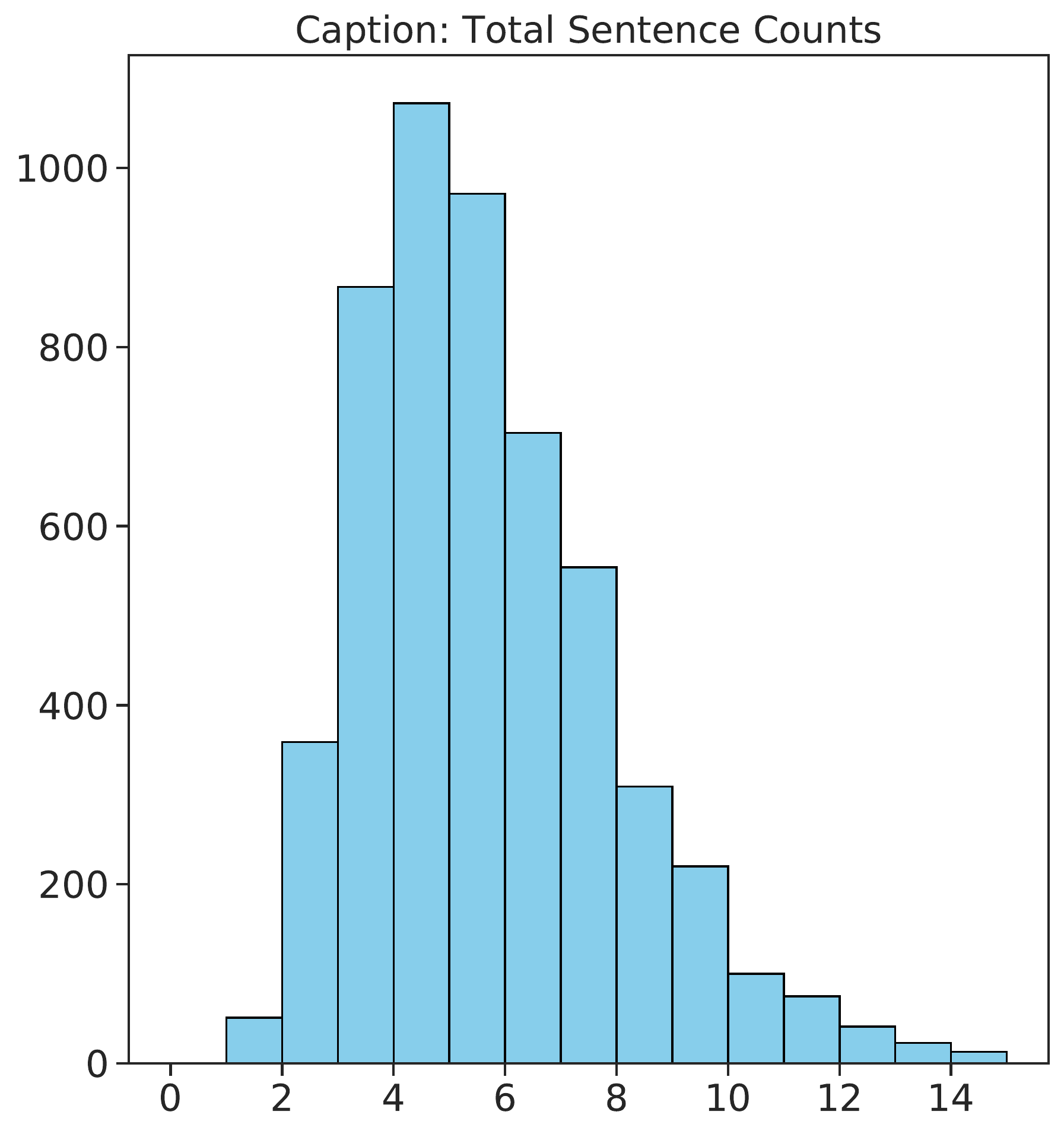}
    \caption{
        Sentence Counts.
    }
    \label{fig:data_wsc_2}
\end{subtable}
\caption{\textbf{Word \& sentence histograms:} the histograms of \textit{per caption} word and sentence counts. The charts can be examined jointly with~\tbref{tab:data-det} for better understandings.}
\label{tab:data_wsc}
\end{figure}

\begin{figure}
\centering
\resizebox{0.8\columnwidth}{!}{%
\begin{tikzpicture}[thick,scale=1, every node/.style={scale=2.2}]
    \coordinate (origin) at (0, 0);
    \foreach[count=\i] \counts/\dim/\color/\rwhere in {
100/protein/black/25,
99/cell/black/25,
88/antibody/black/25,
50/lane/black/24,
46/indicated/black/24,
38/interaction/black/24.5,
36/panel/black/24.5,
33/using/black/24,
32/domain/black/24,
31/control/black/24,
30/complex/black/24,
28/shown/black/24,
28/binding/black/24,
25/used/black/24,
25/extract/black/24,
24/GST/black/24,
22/lysates/black/24,
22/transfected/black/24,
22/incubated/black/24,
22/\shortstack{immuno-\\precipi-\\tated}/black/27,
21/mutant/black/24,
20/analyzed/black/25,
20/assay/black/24,
19/analysis/black/24,
19/Western/black/24,
18/experiment/black/24,
17/bead/black/24,
17/blot/black/24,
17/interacts/black/24,
16/vitro/black/24}
    {
        \coordinate (\i) at (\i * 360 / 30: \counts / 5);
        \node[text=\color] (title) at (\i * 360 / 30: \rwhere) {\Huge\dim};
        \draw[opacity=.5] (origin) -- (title);
    }
    \draw[fill=gray, opacity=0.03] (0, 0) circle [radius=21];
    \draw (0, 0)[opacity=.3, color=blue] circle [radius=20];
    \node[opacity=.8] (title) at (5.4: 20) {\huge100};
    \draw (0, 0)[opacity=.3, color=blue] circle [radius=15];
    \node[opacity=.8] (title) at (6.5: 15) {\huge75};
    \draw (0, 0)[opacity=.3, color=blue] circle [radius=10];
    \node[opacity=.8] (title) at (9.8: 10) {\huge50};
    \draw (0, 0)[opacity=.3, color=blue] circle [radius=5];
    \node (title) at (16: 5) {\huge25};

    \draw [fill=blue!20, opacity=.5] (1) \foreach \i in {2,...,30}{-- (\i)} --cycle;
\end{tikzpicture}%
}
\caption{
    \textbf{Top-30 frequent words:} extracted from all the figure captions in the \dataset~dataset. We normalize the word counts w.r.t the most frequent word, \textit{protein}, which has 4280 appearances (\ie \textit{protein} is denoted as $100\%$). Lemmatization is performed to group words with the same lemmas. 
}
\label{fig:overall_topk_radar_plots}
\end{figure}
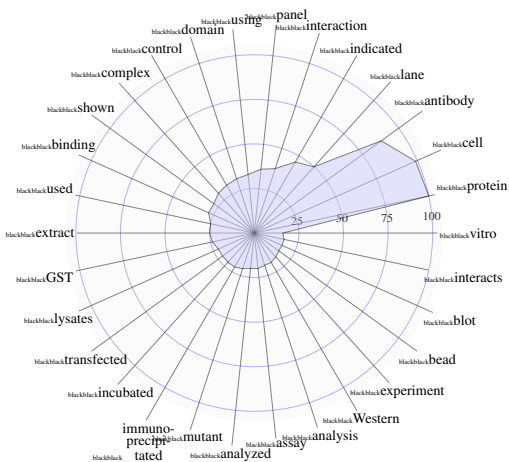

There are in total 5,371 data instances in our dataset, generated from 1,497 OA-PMC articles, with 2,833 uniquely extracted images, as summarized in table~\ref{tab:data-det}.
The total unique label counts of each level in the original IntAct database as well as our collected dataset is summarized in~\tbref{tab:exp-type}.
~\figref{fig:data_wsc_1} shows the histogram of caption word counts of the whole dataset, where the words are tokenized by applying simple NLTK word tokenizer on each caption, and the histogram of sentence counts in a caption is as shown in~\figref{fig:data_wsc_2}.
The top-30 frequent words (stop words and punctuation excluded) of the whole dataset are visualized in~\figref{fig:overall_topk_radar_plots}, with lemmatization applied.

\vspace{0.3em}
\mypar{Data Splits.}
We split the whole dataset into three subsets: train, validation, and test sets, with a ratio of $80\%-10\%-10\%$. In order to prevent models from exploiting certain patterns in the same research article to make predictions during the test time, we assure that no data records extracted from the same paper is split into different subsets,
\ie denote \textit{id} as the paper id from OA-PMC, $\{id | id \in \text{set}_i\} \cap \{id | id \in \text{set}_j\} = \emptyset, i, j \in \{\text{train}, \text{val}, \text{test}\}, i \neq j$. Additionally, we ensure that the labels are distributed evenly in the three sets according to the coarse \textit{participant} method, as illustrated in~\figref{fig:label_dist}.

\section{Benchmark Models}
\label{sec:baselines}

We benchmark several state-of-the-art vision, language and multimodal models against our dataset, that differ primarily by the modalities they encode.
Specifically, we consider unimodal models which take either an image (image-only) or a caption (caption-only) as input, and multimodal models that take both. 
All the output layers for classification are multi-layer perceptrons (MLPs) followed by a softmax layer.


\subsection{Unimodal Models}
\label{ssec:unimodal}

\begin{itemize}[leftmargin=*]
    \item \textbf{Image-Only}: We adopt a variant of convolutional neural networks, ResNet-101~\cite{he2016deep},
    and initialize the networks with two sets of pretrained weights: (1) ImageNet classification task~\cite{deng2009imagenet}, and (2) backbone of Mask R-CNN on object detection task~\cite{he2017mask}.
    We finetune the final three ResNet blocks (from a total of five), given the consistency of early level features across visual domains (more details in the appendix).
    
    \item \textbf{Caption-Only}: We mainly consider the two de-facto variants of language models: LSTM-based~\cite{hochreiter1997long}, and transformer-based~\cite{vaswani2017attention} models. Our LSTM models
    take input word embeddings from Bio-GloVe (300-d)~\cite{burns2019building}. For transformer-based models, we consider two state-of-the-art pretrained masked language models (\textbf{MLM}): BERT~\cite{devlin2019bert} trained on scientific corpora, dubbed SciBERT~\cite{Beltagy2019SciBERT}, and RoBERTa~\cite{liu2019roberta}.
\end{itemize}
We experiment caption-only models with and without the \textbf{masked language finetuning on the caption sentences of our dataset}, by constructing a corpus where each sentence is a caption from the train and validation sets. We use RoBERTa-large and uncased version of SciBERT to initialize the language models' weights.

\begin{figure}[t!]
\centering
    \includegraphics[width=0.75\columnwidth]{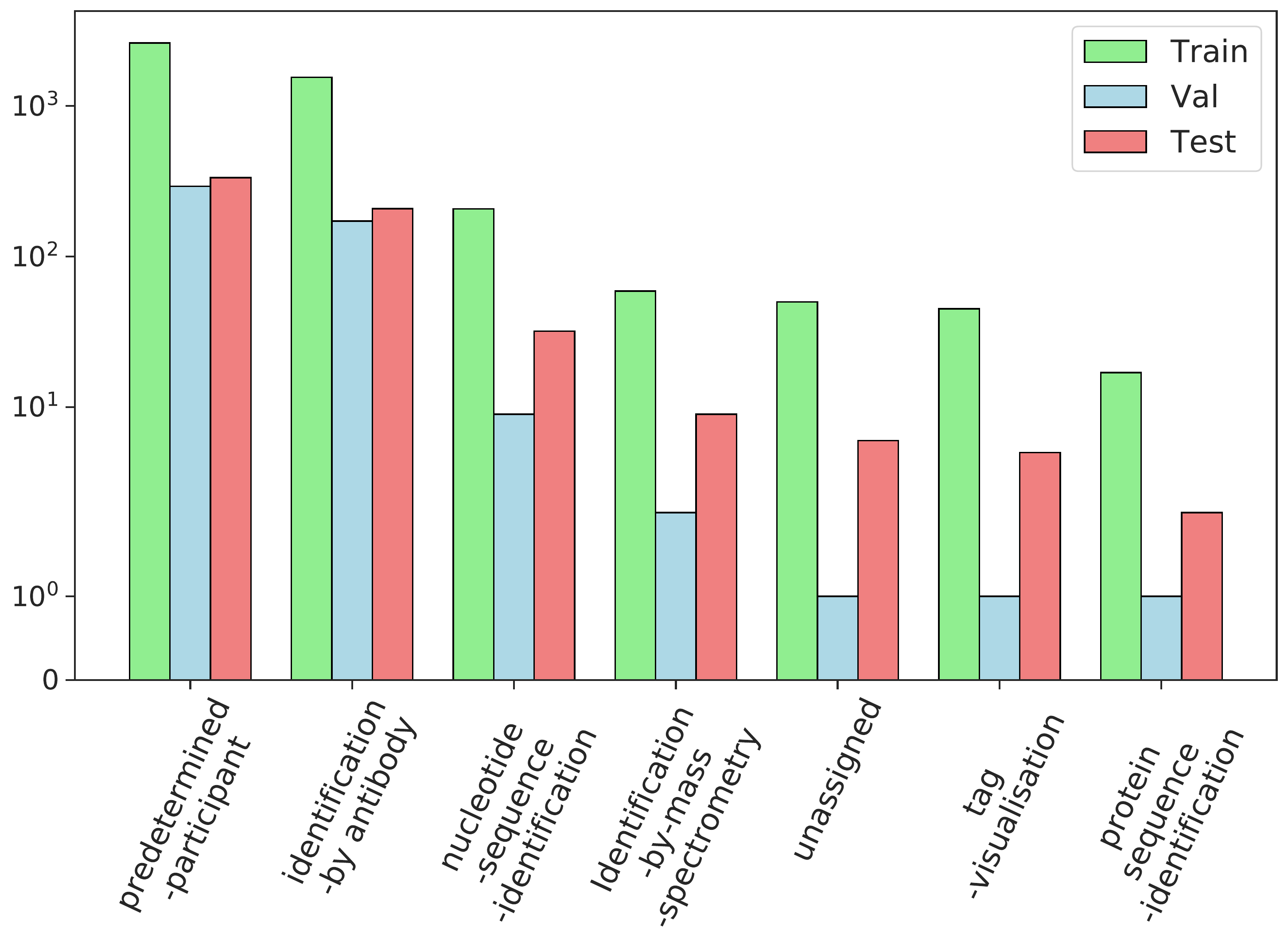}
    \caption{
        \textbf{Label distributions:} with respect to the \textit{participant (coarse)} label type for each data split. We compute the number of data records of each unique ground truth labels. The y-axis is log scaled. The top two classes are: \textit{predetermined participant}, and \textit{identification by antibody}.
    }
    \label{fig:label_dist}
\end{figure}

\newcommand*\colourchecky[1]{%
  \expandafter\newcommand\csname #1checky\endcsname{\ding{51}}%
}
\colourchecky{green}

\newcommand*\colourcross[1]{%
  \expandafter\newcommand\csname #1cross\endcsname{\ding{55}}%
}
\colourcross{red}

\begin{figure}
\begin{subtable}{.47\columnwidth}
\centering
  \includegraphics[width=.9\columnwidth]{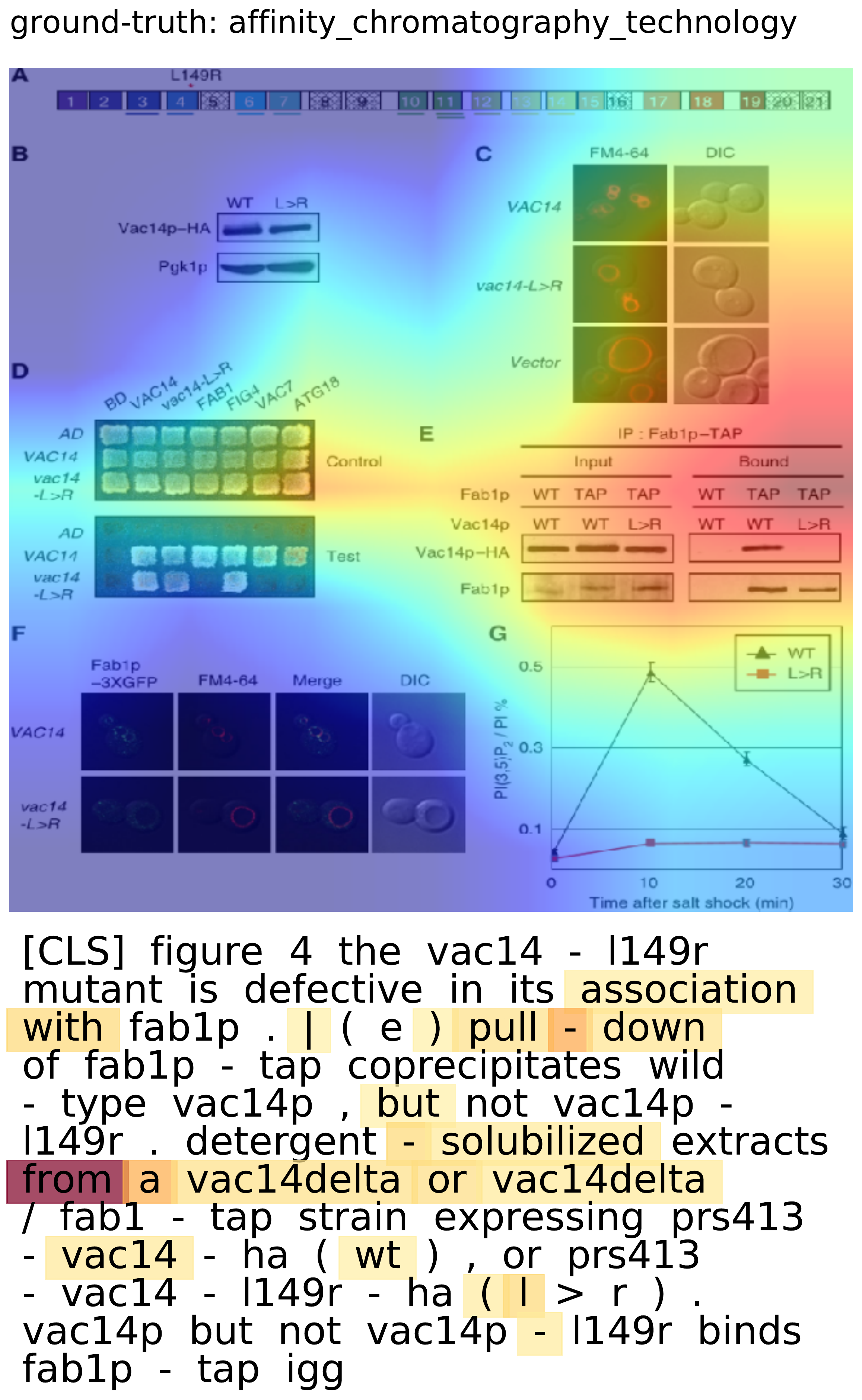}
\end{subtable}%
\quad
\begin{subtable}{.47\columnwidth}
\centering
  \includegraphics[width=.9\columnwidth]{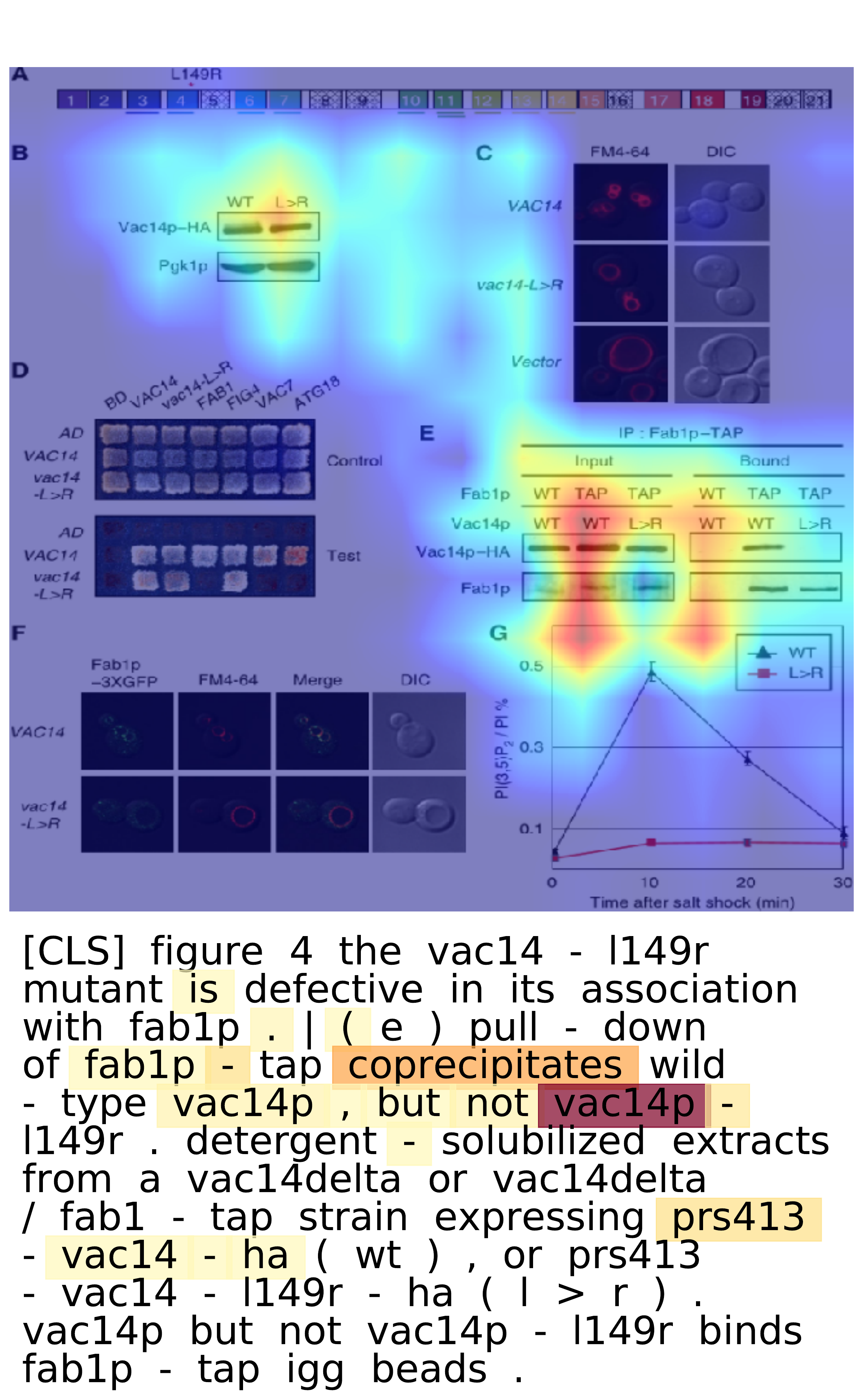}
\end{subtable}

\begin{subtable}{.47\columnwidth}
\centering
  \includegraphics[width=.9\columnwidth]{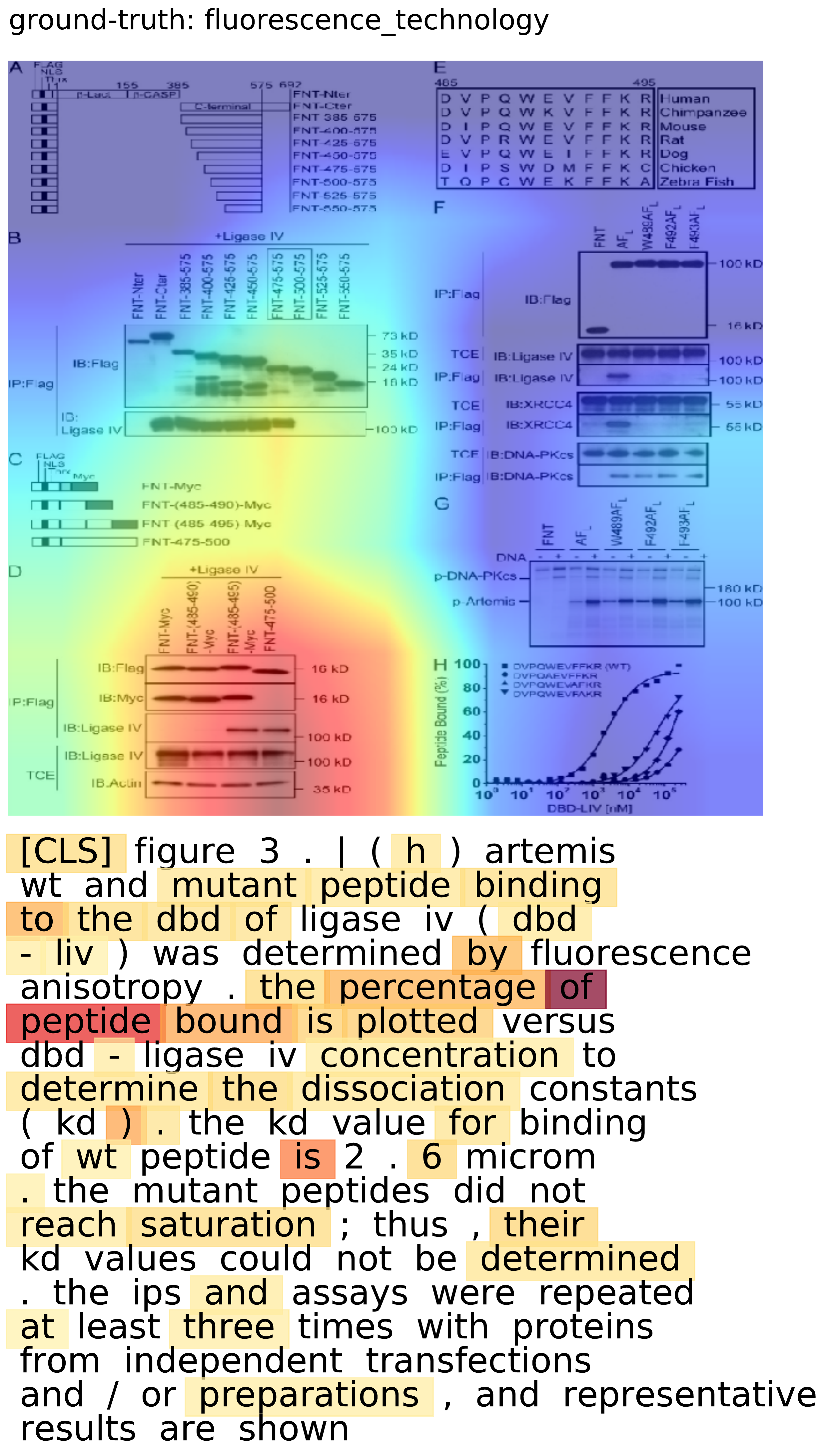}
\end{subtable}%
\quad
\begin{subtable}{.47\columnwidth}
\centering
  \includegraphics[width=.9\columnwidth]{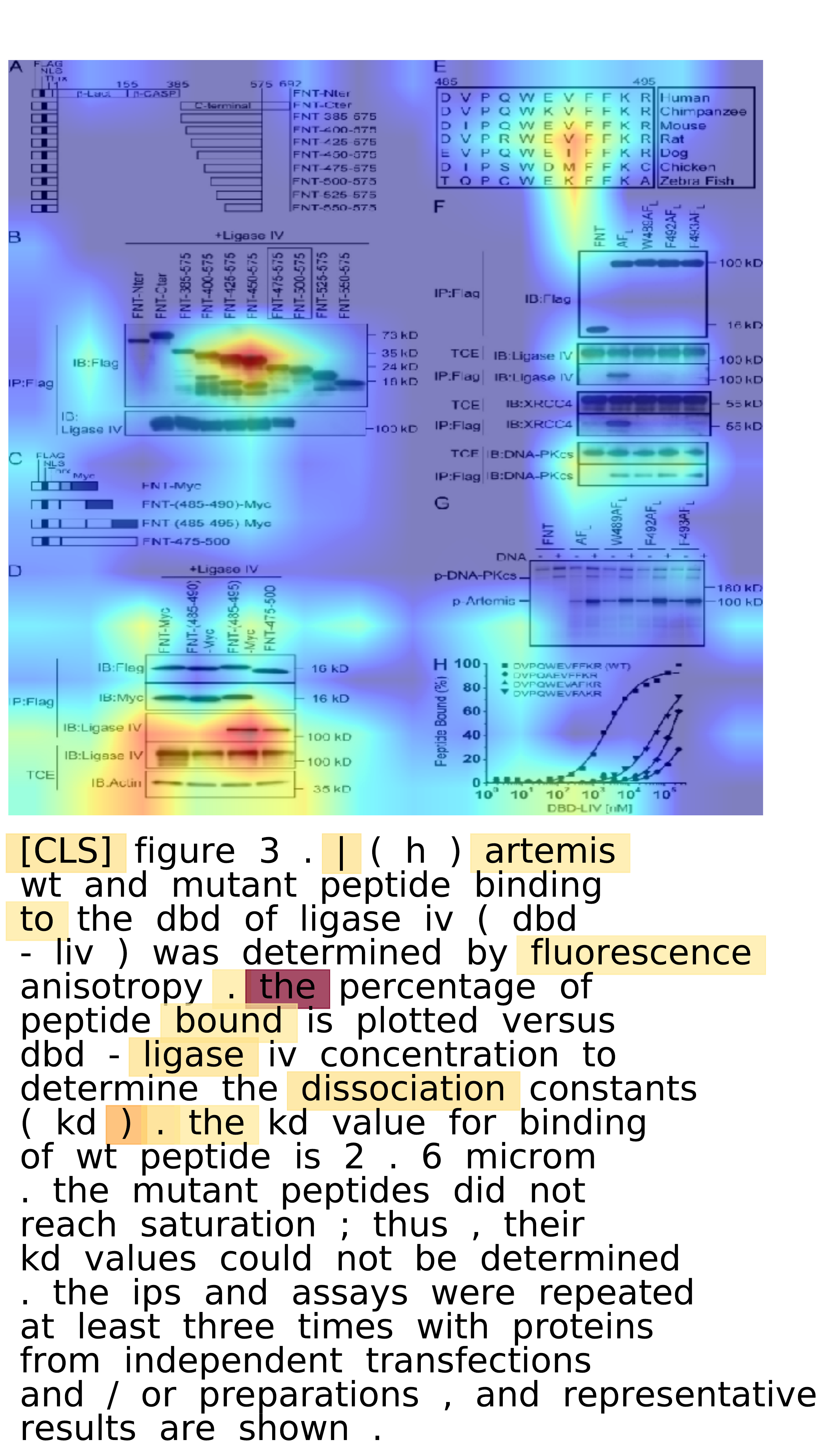}
\end{subtable}

\begin{subtable}{.47\columnwidth}
\centering
  \includegraphics[width=.9\columnwidth]{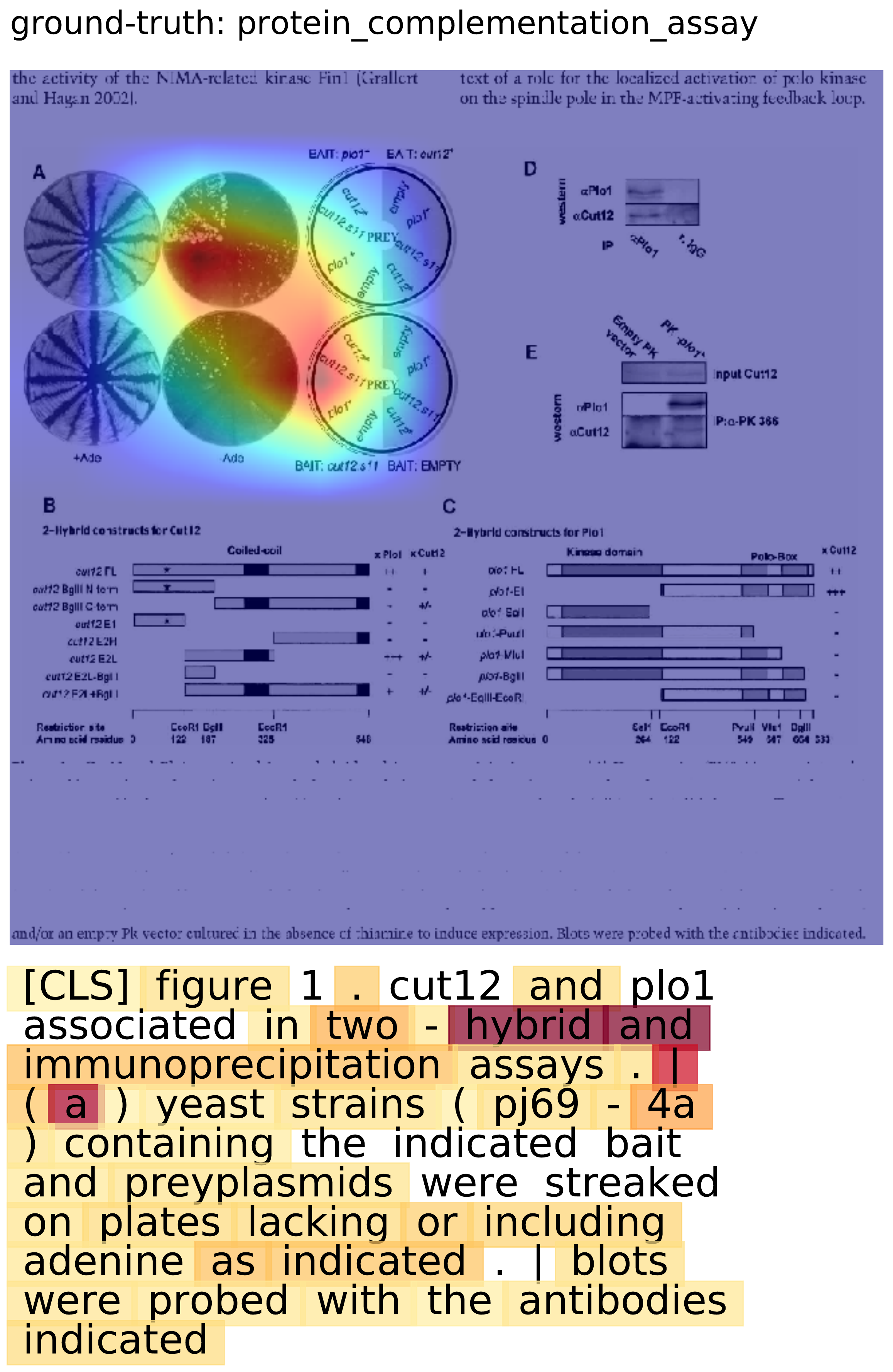}
    \caption{
        Image-only \& Text-only
    }
\end{subtable}%
\quad
\begin{subtable}{.47\columnwidth}
\centering
  \includegraphics[width=.9\columnwidth]{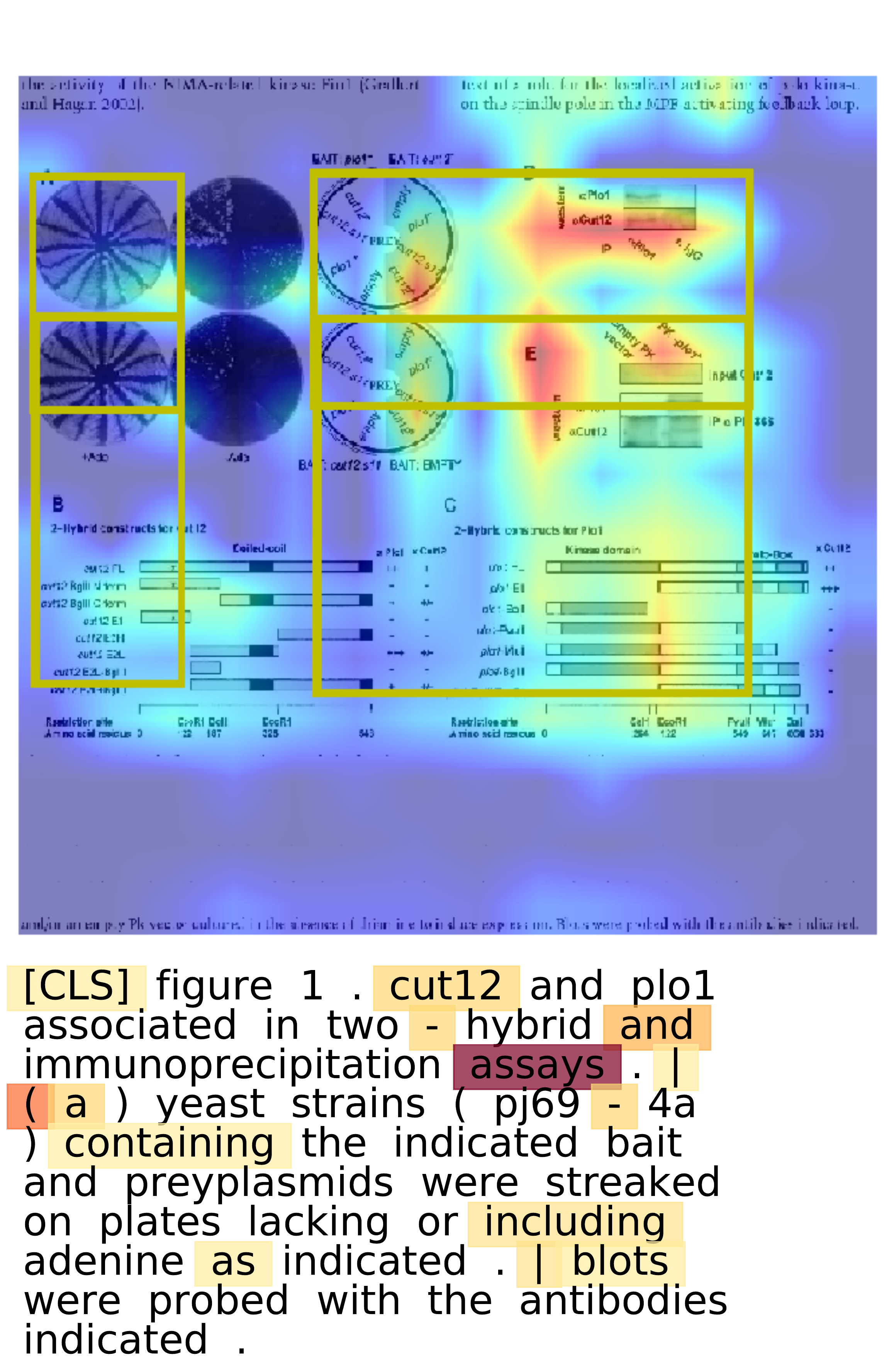}
    \caption{
        Multimodal
    }
\end{subtable}
\caption{\textbf{Saliency Comparisons on Int$_\text{coarse}$} (highest-lowest attention $\rightarrow$ red-blue for images and dark red-light yellow for captions): In each row: \textbf{(a)} \textit{independent} unimodal models -- ResNet-101 \& SciBERT, \textbf{(b)} multimodal model -- VL-BERT. From top to bottom the correctness of predictions between (unimodal, multimodal) is: (\greenchecky, \greenchecky), (\redcross, \greenchecky), and (\greenchecky, \redcross). Unimodal models tend to have more dispersed attentions, while multimodal models have more concentrated salience maps. In the third row, we show top confident generated ROIs on the failure case of multimodal model. 
As shown some of them co-locate with the highest attended regions, which can hypothetically cause the mis-focus.
}
\label{fig:attn_viz}
\end{figure}

\begin{table*}[t!]
\centering
\small
\scalebox{0.95}{
    \begin{tabular}{cclcccccc}
    \toprule
    \textbf{Modalities} & \multicolumn{1}{c}{\textbf{Models}} & \multicolumn{1}{c}{\textbf{Variants}} & \textbf{Par$_{\textbf{coarse}}$} & \textbf{Int$_{\textbf{coarse}}$} & \textbf{Par$_{\textbf{fine}}$} & \textbf{Int$_{\textbf{fine}}$} \\
    \midrule \\[-1.1em]
    
    \multicolumn{1}{c}{---}
    & Majority Baseline
    & \multicolumn{1}{c}{---} & 55.88 & 63.67 & 48.96  & 23.18  \\
    
    \midrule \\[-1.1em]
    \multirow{2}{*}{Image-Only}
    
    & \multirow{2}{*}{ResNet-101}
    & init. from ImageNet & 63.84 & 70.24 & 50.87  & 28.50  \\
    & & init. from MSCoCo & 59.52 & 70.07 & 50.35  & 29.20  \\
    \midrule \\[-1.1em]
    
    \multirow{5}{*}{Caption-Only}
    
    & LSTM w. BioGloVe & \multicolumn{1}{c}{---} & 59.20 & 68.02 & 49.00  & 35.30  \\
    \cline{2-7} \\[-.8em]

    & \multirow{2}{*}{RoBERTa}
    & w/o MLM finetuning & 74.60 & 86.00 & 60.00  & 64.70  \\ 
    & & w. MLM finetuning & 75.40 & 88.60 & 63.00  & 67.10 \\
    \cline{2-7} \\[-.8em]

    & \multirow{2}{*}{SciBERT}
    & w/o MLM finetuning & 76.60 & 86.70 & 62.10  & 65.70  \\ 
    & & w. MLM finetuning & 77.70 & 87.00 & 64.90  & 67.10 \\ 
    \midrule \\[-1.1em]
    
    \multirow{9}{*}{Multi-Modal}

    & \multirow{2}{*}{NLF}
    & w/o language part MLM finetuning & 76.60 & 88.10 & 61.10  & 67.30  \\ 
    & & w. language part MLM finetuning & 73.70 & 87.90 & 62.80  & 70.20 \\
    \cline{2-7} \\[-.8em]

    & \multirow{2}{*}{SAN}
    & w/o language part MLM finetuning & 72.30 & 88.60 & 61.90  & 70.40  \\ 
    & &  w. language part MLM finetuning & 71.60 & 88.90 & 62.80  & 70.40 \\ 
    \cline{2-7} \\[-.8em]
    
    & \multirow{3}{*}{ViL-BERT}
    & w. MLM & 78.20 & 90.64 & 66.26  & 72.15  \\
    & & w. MLM \& NSP & \textbf{78.60} & \textbf{90.83} & 65.57  & 72.84  \\
    & & w. MLM \& NSP \& visual-MLM & 76.47 & 90.48 & 64.19  & 71.80  \\
    \cline{2-7} \\[-.8em]
    
    & \multirow{2}{*}{VL-BERT}
    & w. MLM & 78.02 & 89.96 & \textbf{66.49}  & \textbf{74.65}  \\
    & & w. MLM \& visual-MLM & 77.90 & 89.76 & 65.82 & 74.02 \\
    
    \bottomrule
    \end{tabular}
}
\caption{\textbf{Model accuracies on the test set:} the two label categories are denoted as \textit{Par}, and \textit{Int} for \textit{participant} and \textit{interaction} method respectively. The label hierarchy is indicated as the subscript, \eg Par$_{\textbf{coarse}}$ indicates coarse types of \textit{participant} method. The best performances for each type of labels are bolded, and in all cases the two advanced multimodal models achieve the best performances. Particularly for the two multimodal models, the variants \textbf{without} the \textbf{visual-MLM} objectives perform the best.}
\label{tab:mod-comp}
\end{table*}

\subsection{Multimodal Models}
\label{ssec:multimodal}

\begin{itemize}[leftmargin=*]
    \item \textbf{Naive Late Fusion (NLF)}: The images and captions are encoded by its best performing unimodal models -- ResNet (ImageNet weights) and SciBERT respectively, which are then concatenated (late fusion) and fed into MLPs.
    
    \item \textbf{Stacked Attention Network (SAN)}~\cite{yang2016stacked}: is a multi-step co-attention based framework that has demonstrated good performances on Visual Question Answering (VQA) benchmark~\cite{antol2015vqa}. The image and caption encoders are same as in NLF.

    \item \textbf{ViL-BERT}: Vision-and-Language BERT~\cite{lu2019vilbert}, an extension of BERT model which learns a joint visual-and-linguistics representation through co-attentional transformer layers on top of unimodal visual and textual streams. The model has two major proxy pretraining objectives: (1) textual and visual masked learning, where the visual stream requires the model to predict missing masked-out regions of input images (\textbf{visual-MLM}), and (2) image-text alignment prediction, which extends BERT's next sentence prediction (\textbf{NSP}).
    
    \item \textbf{VL-BERT}: As the concurrent work to ViL-BERT, the visual-linguistics BERT model~\cite{su2019vl} (VL-BERT) performs the multimodal co-attention in an early fusion manner with a single stream of transformer models. VL-BERT also adopts textual and visual masked learning pretraining objectives, while excluding the image-text multimodal alignment prediction.
    
\end{itemize}
The two multimodal BERT models are initialized with the SciBERT pretrained weights directly to their textual parts.
For both ViL-BERT and VL-BERT, the visual-MLM leverages region of interests (ROIs) proposed by the object detection module, as well as the predicted class labels with high confidences.
Due to significant domain shifts between the pretrained object detectors and our dataset, we experiment inclusion and exclusion of various of their proposed pretraining objectives (mainly concerning the visual masked prediction) when \textit{finetuning on our dataset}.
\subsection{Training Details.} 
\label{ssec:setups}

The hyper-parameters for each model are manually tuned against our dataset, and the trained model checkpoints used to evaluate are selected by the best performing ones on the validation set.
All models are trained independently for each method type.
More details of the hyper-parameters and experimental setups can be found in the appendix. 

\section{Experiments and Analysis}
\label{sec:expan}

Our experiments aim to:
(1) Benchmark the performances of the baseline models described in the previous section, and
(2) compare and analyze how and what these models learn.

\vspace{0.3em}
\mypar{Quantitative Results.}
\tbref{tab:mod-comp} summarizes the model performances on the test set, including the majority baseline that selects the most frequent classes in different label types.
All the models, after training on the train set, outperform the majority baseline by large margins, which indicates the sizable training set is effective in transferring knowledge learned from these pretrained models.
The image-only models, despite not having indicators of which sub-figure to look at, still surpass the majority baseline, which we hypothesize that the models still learn the salience in the images to make the correct predictions.
Both transformer-based caption-only models benefit from the masked language finetuning on our dataset, we hypothesize that such finetuning objective can alleviate severe domain shifts between the original pretraining corpora and our~\dataset~corpus.
Among all the models, the two visual-linguistics multimodal models show the best performances on all types of labels, especially on the \textbf{fine-grained types} (\ie \textit{Par$_\text{fine}$} and \textit{Int$_\text{fine}$}). We believe that when granularity is finer, more subtle complementary multimodal understanding is required.
The non-transformer-based multimodal models (NLF and SAN), however, are either on par or worse than the best caption-only models, SciBERT, suggesting that the attention mechanism in transformers may be a substantially better basis for grounding multimodal and complementary information.

The image-only models initialized with classification weights outperforms the one with detection weights, which may hint that the object detectors can be more prone to the \textit{common objects} seen in their original training datasets. Such hypothesis is also shown in the performance comparisons within the visual-linguistics multimodal models, where they tend to perform better without the visual-MLM objective. However, within ViL-BERT, the multimodal alignment objective shown to be beneficial in most label types.
In general, there are still huge gaps between model accuracies and expert human performances (\texttildelow100\% accuracy), especially for the fine-grained types.

\vspace{0.3em}
\mypar{Visualizing What Models Learn.}
We utilize~\textit{Grad-CAM}~\cite{selvaraju2017grad} for visualizing the model salience on the images and~\textit{SmoothGrad}~\cite{smilkov2017smoothgrad} on the captions. 
\figref{fig:attn_viz} shows a sampled side-by-side comparisons between unimodal models (left) and multimodal models (right) of label type~\textit{\textbf{Int$_\text{coarse}$}}. It can be seen that the salience on the images clearly transition from being more dispersed to more detailed and finer-grained from unimodal to multimodal models. Likewise, multimodal models attend less on the common words such as \textit{from}, \textit{and}, \textit{of}, and weight more on domain specific words. The image-only models, without the disambiguation from the captions, tend to focus more on spurious patterns as hinted in the first and second row. While the multimodal models exhibit diverged attentions in the images, it captures the keyword \textit{fluorescence} that the unimodal language model fails to grasp.
The third row of~\figref{fig:attn_viz} shows a failure case of multimodal models, where both unimodal models focused closer to the ideal regions in their inputs (note the sub-figure identifier "\textit{(a)}" in the caption), and hence make the correct predictions. We hypothesize that multimodal models may capture wrong information due to relatively stronger influences by the ROIs proposed by the inherited object detection module (refer to the overlaid yellow-colored ROIs).

\section{Related Works}
\label{sec:relwrk}

\textbf{Multimodal Datasets.}
There are numerous datasets for multimodal machine learning in existence, including visual storytelling~\cite{huang2016visual}, visual-linguistics reasoning~\cite{johnson2017clevr, hasan2019ur,wang2019youmakeup,liu2020violin}, and multimodal question answering (QA)
~\cite{antol2015vqa, tapaswi2016movieqa, kembhavi2016diagram, kembhavi2017you, lei2018tvqa, yagcioglu2018recipeqa, das2018embodied, zellers2019recognition}.
As these works focus on more general domains, our work offers a dataset in the hope of motivating research in domains that often require expertise for labelling, such as biomedical.


\vspace{0.3em}
\mypar{Experiment Method Classification.}
The closest prior work~\cite{burns2019building} has used the figure captions from OA-PMC set to perform similar experiment method classification task. In our \dataset~dataset, we put forth to extract the visual information in conjunctions with the caption texts, and collect a larger-scale dataset.

\vspace{0.3em}
\mypar{Automating Biocuration \& Biomedical Tasks.}
Integrating computational approaches into the workflow of biocuration can be seen in many applications such as constructing genomics knowledge base~\cite{baumgartner2007manual},
biomedical document classification~\cite{cohen2006effective, shatkay2006integrating, jiang2017effective, simon2019bioreader},
biomedical text mining~\cite{dowell2009integrating}, and human-in-the-loop curation~\cite{lee2018scaling}.
Some prior works also adopt multimodal machine learning for general biomedical information extractions~\cite{schlegl2015predicting, ImageCLEFoverview2017, zhang2017tandemnet}, as well as
textual extraction~\cite{burns2017extracting}, medical image captioning~\cite{shin2016learning},
and automated diagnosis from medical images~\cite{jingetal2018automatic, wang2018tienet, liu2019xception}.


Our work aims to further facilitate research in automating biocuration by providing a sizeable multimodal dataset, along with the data collection tool. We benchmark various unimodal and multimodal models with analysis on their strengths that suggest potential improvements.

\section{Conclusions and Future Work}
\label{sec:conc}
In this work, we introduce a new multimodal dataset, \dataset, for biomedical experiment method classification. 
Our dataset comprises extracted image-caption pairs with the associated experiment method labels. As our data is collected in a fully automated \textit{distant supervision} manner, the dataset is easily expandable.

We benchmark the proposed dataset against various baseline models, including state-of-the-art vision models, language models, and multimodal (visual-linguistics) models. The results show that despite multimodal models generally demonstrate superior performances, there are still huge rooms for improvements in the current visual-linguistics grounding paradigms, especially for domain specific data. Hence, we hope this work could motivate the future advancements in multimodal models, primarily on:
(1) low resource domains and better transfer learning.
(2) a less-supervised multimodal grounding method with less reliance on robust pretrained {\em object detectors}.

\section{Acknowledgements}
We thank the anonymous reviewers for their feedback, Yu Hou and Nuan Wen for their quality assessment annotations. This work is supported by a National Institutes of Health (NIH) R01 grant (LM012592). The views and conclusions of this paper are those
of the authors and do not reflect the official policy or position of NIH.

{\fontsize{9.0pt}{10.0pt} \selectfont
\bibliography{aaai2021}

\begin{thebibliography}{57}
\providecommand{\natexlab}[1]{#1}
\providecommand{\url}[1]{\texttt{#1}}
\providecommand{\urlprefix}{URL }
\expandafter\ifx\csname urlstyle\endcsname\relax
  \providecommand{\doi}[1]{doi:\discretionary{}{}{}#1}\else
  \providecommand{\doi}{doi:\discretionary{}{}{}\begingroup
  \urlstyle{rm}\Url}\fi

\bibitem[{Antol et~al.(2015)Antol, Agrawal, Lu, Mitchell, Batra,
  Lawrence~Zitnick, and Parikh}]{antol2015vqa}
Antol, S.; Agrawal, A.; Lu, J.; Mitchell, M.; Batra, D.; Lawrence~Zitnick, C.;
  and Parikh, D. 2015.
\newblock Vqa: Visual question answering.
\newblock In \emph{Proceedings of the IEEE Conference on Computer Vision and
  Pattern Recognition (CVPR)}, 2425--2433.

\bibitem[{Baumgartner~Jr et~al.(2007)Baumgartner~Jr, Cohen, Fox,
  Acquaah-Mensah, and Hunter}]{baumgartner2007manual}
Baumgartner~Jr, W.~A.; Cohen, K.~B.; Fox, L.~M.; Acquaah-Mensah, G.; and
  Hunter, L. 2007.
\newblock Manual curation is not sufficient for annotation of genomic
  databases.
\newblock In \emph{Bioinformatics}, volume~23, i41--i48. Oxford University
  Press.

\bibitem[{Beltagy, Lo, and Cohan(2019)}]{Beltagy2019SciBERT}
Beltagy, I.; Lo, K.; and Cohan, A. 2019.
\newblock SciBERT: Pretrained Language Model for Scientific Text.
\newblock In \emph{Empirical Methods in Natural Language Processing (EMNLP)}.

\bibitem[{{Ben Abacha} et~al.(2019){Ben Abacha}, Hasan, Datla, Liu,
  Demner-Fushman, and M\"uller}]{ImageCLEFVQA-Med2019}
{Ben Abacha}, A.; Hasan, S.~A.; Datla, V.~V.; Liu, J.; Demner-Fushman, D.; and
  M\"uller, H. 2019.
\newblock {VQA-Med}: Overview of the Medical Visual Question Answering Task at
  ImageCLEF 2019.
\newblock In \emph{CLEF2019 Working Notes}, {CEUR} Workshop Proceedings.
  Lugano, Switzerland: CEUR-WS.org $<$http://ceur-ws.org$>$.

\bibitem[{Burns et~al.(2018)Burns, Shi, Wu, Cao, and
  Natarajan}]{burns2018towards}
Burns, G.; Shi, X.; Wu, Y.; Cao, H.; and Natarajan, P. 2018.
\newblock Towards Evidence Extraction: Analysis of Scientific Figures from
  Studies of Molecular Interactions.
\newblock In \emph{ISWC (Best Workshop Papers)}, 95--102.

\bibitem[{Burns, Dasigi, and Hovy(2017)}]{burns2017extracting}
Burns, G.~A.; Dasigi, P.; and Hovy, E.~H. 2017.
\newblock Extracting evidence fragments for distant supervision of molecular
  interactions.
\newblock In \emph{BioRxiv}, 192856. Cold Spring Harbor Laboratory.

\bibitem[{Burns, Li, and Peng(2019)}]{burns2019building}
Burns, G.~A.; Li, X.; and Peng, N. 2019.
\newblock Building deep learning models for evidence classification from the
  open access biomedical literature.
\newblock In \emph{Database}, volume 2019. Narnia.

\bibitem[{Chen et~al.(2020)Chen, Li, Yu, Kholy, Ahmed, Gan, Cheng, and
  Liu}]{chen2019uniter}
Chen, Y.-C.; Li, L.; Yu, L.; Kholy, A.~E.; Ahmed, F.; Gan, Z.; Cheng, Y.; and
  Liu, J. 2020.
\newblock Uniter: Learning universal image-text representations.
\newblock In \emph{European Conference on Computer Vision (ECCV)}.

\bibitem[{Cohen(2006)}]{cohen2006effective}
Cohen, A.~M. 2006.
\newblock An effective general purpose approach for automated biomedical
  document classification.
\newblock In \emph{AMIA annual symposium proceedings}, volume 2006, 161.
  American Medical Informatics Association.

\bibitem[{Craven, Kumlien et~al.(1999)}]{craven1999constructing}
Craven, M.; Kumlien, J.; et~al. 1999.
\newblock Constructing biological knowledge bases by extracting information
  from text sources.
\newblock In \emph{ISMB}, volume 1999, 77--86.

\bibitem[{Das et~al.(2018)Das, Datta, Gkioxari, Lee, Parikh, and
  Batra}]{das2018embodied}
Das, A.; Datta, S.; Gkioxari, G.; Lee, S.; Parikh, D.; and Batra, D. 2018.
\newblock Embodied question answering.
\newblock In \emph{Proceedings of the IEEE Conference on Computer Vision and
  Pattern Recognition (CVPR)}, 2054--2063.

\bibitem[{Demner-Fushman et~al.(2012)Demner-Fushman, Antani, Simpson, and
  Thoma}]{demner2012design}
Demner-Fushman, D.; Antani, S.; Simpson, M.; and Thoma, G.~R. 2012.
\newblock Design and development of a multimodal biomedical information
  retrieval system.
\newblock In \emph{Journal of Computing Science and Engineering}, volume~6,
  168--177. Korean Institute of Information Scientists and Engineers.

\bibitem[{Deng et~al.(2009)Deng, Dong, Socher, Li, Li, and
  Fei-Fei}]{deng2009imagenet}
Deng, J.; Dong, W.; Socher, R.; Li, L.-J.; Li, K.; and Fei-Fei, L. 2009.
\newblock Imagenet: A large-scale hierarchical image database.
\newblock In \emph{Proceedings of the IEEE Conference on Computer Vision and
  Pattern Recognition (CVPR)}, 248--255.

\bibitem[{Devlin et~al.(2019)Devlin, Chang, Lee, and
  Toutanova}]{devlin2019bert}
Devlin, J.; Chang, M.-W.; Lee, K.; and Toutanova, K. 2019.
\newblock BERT: Pre-training of Deep Bidirectional Transformers for Language
  Understanding.
\newblock In \emph{North American Chapter of the Association for Computational
  Linguistics (NAACL-HLT)}, 4171--4186.

\bibitem[{Dowell et~al.(2009)Dowell, McAndrews-Hill, Hill, Drabkin, and
  Blake}]{dowell2009integrating}
Dowell, K.~G.; McAndrews-Hill, M.~S.; Hill, D.~P.; Drabkin, H.~J.; and Blake,
  J.~A. 2009.
\newblock Integrating text mining into the MGI biocuration workflow.
\newblock In \emph{Database}, volume 2009. Narnia.

\bibitem[{Eickhoff et~al.(2017)Eickhoff, Schwall, Garc\'ia Seco~de Herrera, and
  M\"uller}]{ImageCLEFoverview2017}
Eickhoff, C.; Schwall, I.; Garc\'ia Seco~de Herrera, A.; and M\"uller, H. 2017.
\newblock Overview of {ImageCLEFcaption} 2017 - the Image Caption Prediction
  and Concept Extraction Tasks to Understand Biomedical Images.
\newblock In \emph{CLEF (Working Notes)}, {CEUR} Workshop Proceedings. Dublin,
  Ireland: CEUR-WS.org $<$http://ceur-ws.org$>$.

\bibitem[{Gururangan et~al.(2020)Gururangan, Marasovi{\'c}, Swayamdipta, Lo,
  Beltagy, Downey, and Smith}]{gururangan2020don}
Gururangan, S.; Marasovi{\'c}, A.; Swayamdipta, S.; Lo, K.; Beltagy, I.;
  Downey, D.; and Smith, N.~A. 2020.
\newblock Don't Stop Pretraining: Adapt Language Models to Domains and Tasks.
\newblock \emph{arXiv preprint arXiv:2004.10964} .

\bibitem[{Hasan et~al.(2019)Hasan, Rahman, Zadeh, Zhong, Tanveer, Morency
  et~al.}]{hasan2019ur}
Hasan, M.~K.; Rahman, W.; Zadeh, A.; Zhong, J.; Tanveer, M.~I.; Morency, L.-P.;
  et~al. 2019.
\newblock UR-FUNNY: A multimodal language dataset for understanding humor.
\newblock In \emph{Empirical Methods in Natural Language Processing (EMNLP)}.

\bibitem[{He et~al.(2017)He, Gkioxari, Doll{\'a}r, and Girshick}]{he2017mask}
He, K.; Gkioxari, G.; Doll{\'a}r, P.; and Girshick, R. 2017.
\newblock Mask r-cnn.
\newblock In \emph{International Conference on Computer Vision (ICCV)},
  2961--2969.

\bibitem[{He et~al.(2016)He, Zhang, Ren, and Sun}]{he2016deep}
He, K.; Zhang, X.; Ren, S.; and Sun, J. 2016.
\newblock Deep residual learning for image recognition.
\newblock In \emph{Proceedings of the IEEE Conference on Computer Vision and
  Pattern Recognition (CVPR)}, 770--778.

\bibitem[{He et~al.(2020)He, Zhang, Mou, Xing, and Xie}]{he2020pathvqa}
He, X.; Zhang, Y.; Mou, L.; Xing, E.; and Xie, P. 2020.
\newblock PathVQA: 30000+ Questions for Medical Visual Question Answering.
\newblock \emph{arXiv preprint arXiv:2003.10286} .

\bibitem[{Hochreiter and Schmidhuber(1997)}]{hochreiter1997long}
Hochreiter, S.; and Schmidhuber, J. 1997.
\newblock Long short-term memory.
\newblock In \emph{Neural computation}, volume~9, 1735--1780. MIT Press.

\bibitem[{Huang et~al.(2016)Huang, Ferraro, Mostafazadeh, Misra, Agrawal,
  Devlin, Girshick, He, Kohli, Batra et~al.}]{huang2016visual}
Huang, T.-H.; Ferraro, F.; Mostafazadeh, N.; Misra, I.; Agrawal, A.; Devlin,
  J.; Girshick, R.; He, X.; Kohli, P.; Batra, D.; et~al. 2016.
\newblock Visual storytelling.
\newblock In \emph{North American Chapter of the Association for Computational
  Linguistics (NAACL-HLT)}, 1233--1239.

\bibitem[{ISB(2018)}]{10.1371/journal.pbio.2002846}
ISB. 2018.
\newblock Biocuration: Distilling data into knowledge.
\newblock In \emph{PLOS Biology}, volume~16, 1--8. Public Library of Science.
\newblock \doi{10.1371/journal.pbio.2002846}.
\newblock \urlprefix\url{https://doi.org/10.1371/journal.pbio.2002846}.

\bibitem[{Jiang et~al.(2017)Jiang, Ringwald, Blake, and
  Shatkay}]{jiang2017effective}
Jiang, X.; Ringwald, M.; Blake, J.; and Shatkay, H. 2017.
\newblock Effective biomedical document classification for identifying
  publications relevant to the mouse Gene Expression Database (GXD).
\newblock In \emph{Database}, volume 2017. Narnia.

\bibitem[{Jing, Xie, and Xing(2018)}]{jingetal2018automatic}
Jing, B.; Xie, P.; and Xing, E. 2018.
\newblock On the Automatic Generation of Medical Imaging Reports.
\newblock In \emph{Association for Computational Linguistics (ACL)},
  2577--2586. Melbourne, Australia: Association for Computational Linguistics.
\newblock \doi{10.18653/v1/P18-1240}.

\bibitem[{Johnson et~al.(2017)Johnson, Hariharan, van~der Maaten, Fei-Fei,
  Lawrence~Zitnick, and Girshick}]{johnson2017clevr}
Johnson, J.; Hariharan, B.; van~der Maaten, L.; Fei-Fei, L.; Lawrence~Zitnick,
  C.; and Girshick, R. 2017.
\newblock Clevr: A diagnostic dataset for compositional language and elementary
  visual reasoning.
\newblock In \emph{Proceedings of the IEEE Conference on Computer Vision and
  Pattern Recognition (CVPR)}, 2901--2910.

\bibitem[{Kembhavi et~al.(2016)Kembhavi, Salvato, Kolve, Seo, Hajishirzi, and
  Farhadi}]{kembhavi2016diagram}
Kembhavi, A.; Salvato, M.; Kolve, E.; Seo, M.; Hajishirzi, H.; and Farhadi, A.
  2016.
\newblock A diagram is worth a dozen images.
\newblock In \emph{European Conference on Computer Vision (ECCV)}, 235--251.
  Springer.

\bibitem[{Kembhavi et~al.(2017)Kembhavi, Seo, Schwenk, Choi, Farhadi, and
  Hajishirzi}]{kembhavi2017you}
Kembhavi, A.; Seo, M.; Schwenk, D.; Choi, J.; Farhadi, A.; and Hajishirzi, H.
  2017.
\newblock Are you smarter than a sixth grader? textbook question answering for
  multimodal machine comprehension.
\newblock In \emph{Proceedings of the IEEE Conference on Computer Vision and
  Pattern Recognition (CVPR)}, 4999--5007.

\bibitem[{Krishna et~al.(2017)Krishna, Zhu, Groth, Johnson, Hata, Kravitz,
  Chen, Kalantidis, Li, Shamma et~al.}]{krishna2017visual}
Krishna, R.; Zhu, Y.; Groth, O.; Johnson, J.; Hata, K.; Kravitz, J.; Chen, S.;
  Kalantidis, Y.; Li, L.-J.; Shamma, D.~A.; et~al. 2017.
\newblock Visual genome: Connecting language and vision using crowdsourced
  dense image annotations.
\newblock \emph{International Journal of Computer Vision (IJCV)} 123(1):
  32--73.

\bibitem[{Lee et~al.(2018)Lee, Famiglietti, McMahon, Wei, MacArthur, Poux,
  Breuza, Bridge, Cunningham, Xenarios et~al.}]{lee2018scaling}
Lee, K.; Famiglietti, M.~L.; McMahon, A.; Wei, C.-H.; MacArthur, J. A.~L.;
  Poux, S.; Breuza, L.; Bridge, A.; Cunningham, F.; Xenarios, I.; et~al. 2018.
\newblock Scaling up data curation using deep learning: An application to
  literature triage in genomic variation resources.
\newblock In \emph{PLoS computational biology}, volume~14, e1006390. Public
  Library of Science.

\bibitem[{Lei et~al.(2018)Lei, Yu, Bansal, and Berg}]{lei2018tvqa}
Lei, J.; Yu, L.; Bansal, M.; and Berg, T.~L. 2018.
\newblock Tvqa: Localized, compositional video question answering.
\newblock In \emph{Empirical Methods in Natural Language Processing (EMNLP)}.

\bibitem[{Li et~al.(2019)Li, Yatskar, Yin, Hsieh, and Chang}]{li2019visualbert}
Li, L.~H.; Yatskar, M.; Yin, D.; Hsieh, C.-J.; and Chang, K.-W. 2019.
\newblock Visualbert: A simple and performant baseline for vision and language.
\newblock \emph{arXiv preprint arXiv:1908.03557} .

\bibitem[{Liu et~al.(2020)Liu, Chen, Cheng, Gan, Yu, Yang, and
  Liu}]{liu2020violin}
Liu, J.; Chen, W.; Cheng, Y.; Gan, Z.; Yu, L.; Yang, Y.; and Liu, J. 2020.
\newblock VIOLIN: A Large-Scale Dataset for Video-and-Language Inference.
\newblock In \emph{Proceedings of the IEEE Conference on Computer Vision and
  Pattern Recognition (CVPR)}.

\bibitem[{Liu et~al.(2019{\natexlab{a}})Liu, Ou, Che, Zhou, and
  Ding}]{liu2019xception}
Liu, S.; Ou, X.; Che, J.; Zhou, X.; and Ding, H. 2019{\natexlab{a}}.
\newblock An Xception-GRU Model for Visual Question Answering in the Medical
  Domain.
\newblock \emph{CLEF (Working Notes)} .

\bibitem[{Liu et~al.(2019{\natexlab{b}})Liu, Ott, Goyal, Du, Joshi, Chen, Levy,
  Lewis, Zettlemoyer, and Stoyanov}]{liu2019roberta}
Liu, Y.; Ott, M.; Goyal, N.; Du, J.; Joshi, M.; Chen, D.; Levy, O.; Lewis, M.;
  Zettlemoyer, L.; and Stoyanov, V. 2019{\natexlab{b}}.
\newblock Roberta: A robustly optimized bert pretraining approach.
\newblock \emph{arXiv preprint arXiv:1907.11692} .

\bibitem[{Lu et~al.(2019)Lu, Batra, Parikh, and Lee}]{lu2019vilbert}
Lu, J.; Batra, D.; Parikh, D.; and Lee, S. 2019.
\newblock Vilbert: Pretraining task-agnostic visiolinguistic representations
  for vision-and-language tasks.
\newblock In \emph{Advances in Neural Information Processing Systems
  (NeurIPS)}, 13--23.

\bibitem[{Mintz et~al.(2009)Mintz, Bills, Snow, and
  Jurafsky}]{mintz2009distant}
Mintz, M.; Bills, S.; Snow, R.; and Jurafsky, D. 2009.
\newblock Distant supervision for relation extraction without labeled data.
\newblock In \emph{Proceedings of the Joint Conference of the 47th Annual
  Meeting of the ACL and the 4th International Joint Conference on Natural
  Language Processing of the AFNLP}, 1003--1011.

\bibitem[{Mohan et~al.(2018)Mohan, Fiorini, Kim, and Lu}]{mohan2018fast}
Mohan, S.; Fiorini, N.; Kim, S.; and Lu, Z. 2018.
\newblock A fast deep learning model for textual relevance in biomedical
  information retrieval.
\newblock In \emph{Proceedings of the 2018 World Wide Web Conference}, 77--86.

\bibitem[{Nguyen et~al.(2019)Nguyen, Do, Nguyen, Do, Tjiputra, and
  Tran}]{nguyen2019overcoming}
Nguyen, B.~D.; Do, T.-T.; Nguyen, B.~X.; Do, T.; Tjiputra, E.; and Tran, Q.~D.
  2019.
\newblock Overcoming Data Limitation in Medical Visual Question Answering.
\newblock In \emph{International Conference on Medical Image Computing and
  Computer-Assisted Intervention (MICCAI)}. Springer.

\bibitem[{Orchard et~al.(2013)Orchard, Ammari, Aranda, Breuza, Briganti,
  Broackes-Carter, Campbell, Chavali, Chen, Del-Toro
  et~al.}]{orchard2013mintact}
Orchard, S.; Ammari, M.; Aranda, B.; Breuza, L.; Briganti, L.; Broackes-Carter,
  F.; Campbell, N.~H.; Chavali, G.; Chen, C.; Del-Toro, N.; et~al. 2013.
\newblock The MIntAct project—IntAct as a common curation platform for 11
  molecular interaction databases.
\newblock \emph{Nucleic Acids Research (NAR)} 42(D1): D358--D363.

\bibitem[{Schlegl et~al.(2015)Schlegl, Waldstein, Vogl, Schmidt-Erfurth, and
  Langs}]{schlegl2015predicting}
Schlegl, T.; Waldstein, S.~M.; Vogl, W.-D.; Schmidt-Erfurth, U.; and Langs, G.
  2015.
\newblock Predicting semantic descriptions from medical images with
  convolutional neural networks.
\newblock In \emph{International Conference on Information Processing in
  Medical Imaging (IPMI)}, 437--448. Springer.

\bibitem[{Selvaraju et~al.(2017)Selvaraju, Cogswell, Das, Vedantam, Parikh, and
  Batra}]{selvaraju2017grad}
Selvaraju, R.~R.; Cogswell, M.; Das, A.; Vedantam, R.; Parikh, D.; and Batra,
  D. 2017.
\newblock Grad-cam: Visual explanations from deep networks via gradient-based
  localization.
\newblock In \emph{International Conference on Computer Vision (ICCV)},
  618--626.

\bibitem[{Shatkay, Chen, and Blostein(2006)}]{shatkay2006integrating}
Shatkay, H.; Chen, N.; and Blostein, D. 2006.
\newblock Integrating image data into biomedical text categorization.
\newblock In \emph{Bioinformatics}, volume~22, e446--e453. Oxford University
  Press.

\bibitem[{Shin et~al.(2016)Shin, Roberts, Lu, Demner-Fushman, Yao, and
  Summers}]{shin2016learning}
Shin, H.-C.; Roberts, K.; Lu, L.; Demner-Fushman, D.; Yao, J.; and Summers,
  R.~M. 2016.
\newblock Learning to read chest x-rays: Recurrent neural cascade model for
  automated image annotation.
\newblock In \emph{Proceedings of the IEEE Conference on Computer Vision and
  Pattern Recognition (CVPR)}, 2497--2506.

\bibitem[{Simon et~al.(2019)Simon, Davidsen, Hansen, Seymour, Barnkob, and
  Olsen}]{simon2019bioreader}
Simon, C.; Davidsen, K.; Hansen, C.; Seymour, E.; Barnkob, M.~B.; and Olsen,
  L.~R. 2019.
\newblock BioReader: a text mining tool for performing classification of
  biomedical literature.
\newblock In \emph{BMC bioinformatics}, volume~19, 57. Springer.

\bibitem[{Smilkov et~al.(2017)Smilkov, Thorat, Kim, Vi{\'e}gas, and
  Wattenberg}]{smilkov2017smoothgrad}
Smilkov, D.; Thorat, N.; Kim, B.; Vi{\'e}gas, F.; and Wattenberg, M. 2017.
\newblock Smoothgrad: removing noise by adding noise.
\newblock In \emph{International Conference on Machine Learning (ICML)}.
  Workshop on Visualization for Deep Learning.

\bibitem[{Su et~al.(2020)Su, Zhu, Cao, Li, Lu, Wei, and Dai}]{su2019vl}
Su, W.; Zhu, X.; Cao, Y.; Li, B.; Lu, L.; Wei, F.; and Dai, J. 2020.
\newblock Vl-bert: Pre-training of generic visual-linguistic representations.
\newblock In \emph{International Conference on Learning Representations
  (ICLR)}.

\bibitem[{Tapaswi et~al.(2016)Tapaswi, Zhu, Stiefelhagen, Torralba, Urtasun,
  and Fidler}]{tapaswi2016movieqa}
Tapaswi, M.; Zhu, Y.; Stiefelhagen, R.; Torralba, A.; Urtasun, R.; and Fidler,
  S. 2016.
\newblock Movieqa: Understanding stories in movies through question-answering.
\newblock In \emph{Proceedings of the IEEE Conference on Computer Vision and
  Pattern Recognition (CVPR)}, 4631--4640.

\bibitem[{Vaswani et~al.(2017)Vaswani, Shazeer, Parmar, Uszkoreit, Jones,
  Gomez, Kaiser, and Polosukhin}]{vaswani2017attention}
Vaswani, A.; Shazeer, N.; Parmar, N.; Uszkoreit, J.; Jones, L.; Gomez, A.~N.;
  Kaiser, {\L}.; and Polosukhin, I. 2017.
\newblock Attention is all you need.
\newblock In \emph{Advances in Neural Information Processing Systems
  (NeurIPS)}, 5998--6008.

\bibitem[{Wang et~al.(2019)Wang, Wang, Chen, and Jin}]{wang2019youmakeup}
Wang, W.; Wang, Y.; Chen, S.; and Jin, Q. 2019.
\newblock YouMakeup: A Large-Scale Domain-Specific Multimodal Dataset for
  Fine-Grained Semantic Comprehension.
\newblock In \emph{Proceedings of the 2019 Conference on Empirical Methods in
  Natural Language Processing and the 9th International Joint Conference on
  Natural Language Processing (EMNLP-IJCNLP)}, 5136--5146.

\bibitem[{Wang et~al.(2018)Wang, Peng, Lu, Lu, and Summers}]{wang2018tienet}
Wang, X.; Peng, Y.; Lu, L.; Lu, Z.; and Summers, R.~M. 2018.
\newblock Tienet: Text-image embedding network for common thorax disease
  classification and reporting in chest x-rays.
\newblock In \emph{Proceedings of the IEEE Conference on Computer Vision and
  Pattern Recognition (CVPR)}, 9049--9058.

\bibitem[{Yagcioglu et~al.(2018)Yagcioglu, Erdem, Erdem, and
  Ikizler-Cinbis}]{yagcioglu2018recipeqa}
Yagcioglu, S.; Erdem, A.; Erdem, E.; and Ikizler-Cinbis, N. 2018.
\newblock Recipeqa: A challenge dataset for multimodal comprehension of cooking
  recipes.
\newblock In \emph{Empirical Methods in Natural Language Processing (EMNLP)}.

\bibitem[{Yang et~al.(2016)Yang, He, Gao, Deng, and Smola}]{yang2016stacked}
Yang, Z.; He, X.; Gao, J.; Deng, L.; and Smola, A. 2016.
\newblock Stacked attention networks for image question answering.
\newblock In \emph{Proceedings of the IEEE Conference on Computer Vision and
  Pattern Recognition (CVPR)}, 21--29.

\bibitem[{Yosinski et~al.(2014)Yosinski, Clune, Bengio, and
  Lipson}]{yosinski2014transferable}
Yosinski, J.; Clune, J.; Bengio, Y.; and Lipson, H. 2014.
\newblock How transferable are features in deep neural networks?
\newblock In \emph{Advances in Neural Information Processing Systems
  (NeurIPS)}, 3320--3328.

\bibitem[{Zellers et~al.(2019)Zellers, Bisk, Farhadi, and
  Choi}]{zellers2019recognition}
Zellers, R.; Bisk, Y.; Farhadi, A.; and Choi, Y. 2019.
\newblock From recognition to cognition: Visual commonsense reasoning.
\newblock In \emph{Proceedings of the IEEE Conference on Computer Vision and
  Pattern Recognition (CVPR)}, 6720--6731.

\bibitem[{Zhang et~al.(2017)Zhang, Chen, Sapkota, and
  Yang}]{zhang2017tandemnet}
Zhang, Z.; Chen, P.; Sapkota, M.; and Yang, L. 2017.
\newblock Tandemnet: Distilling knowledge from medical images using diagnostic
  reports as optional semantic references.
\newblock In \emph{International Conference on Medical Image Computing and
  Computer-Assisted Intervention (MICCAI)}, 320--328. Springer.

\end{thebibliography}
}

\clearpage

\appendix

\section{More on The \dataset~Dataset}
\label{a-sec:dataset}

\subsection{All Label Types}
\label{a-sec:all_labels}

The descriptions of the experiment method labels featured in our \dataset~dataset, across all levels of the two categories, can be referred to from the original IntAct database\footnote{https://www.ebi.ac.uk/ols/ontologies/mi}, specifically for \textit{participant identification}\footnote{\url{https://www.ebi.ac.uk/ols/ontologies/mi/terms?iri=http\%3A\%2F\%2Fpurl.obolibrary.org\%2Fobo\%2FMI_0002\&viewMode=All&siblings=false}} method types, and \textit{interaction detection}\footnote{\url{https://www.ebi.ac.uk/ols/ontologies/mi/terms?iri=http\%3A\%2F\%2Fpurl.obolibrary.org\%2Fobo\%2FMI_0001\&viewMode=All\&siblings=false}} method types. In our full release of the \dataset~dataset, we also include a \texttt{.tsv} file for the hierarchical mappings among the coarse and fine label types.

\subsection{Details of Sub-Caption Segmentation}
\label{a-ssec: sub_cap_seg}

\paragraph{Segmentation}
In general, we try to segment the sub-captions as systematically as possible. There are compound sub-figure identifiers that can refer to multiple sub-figures at once, \eg "(A,C)" or \eg "(A-D)" will represent sub-figure "A" and "C", and sub-figures "A" to "D", respectively. In these cases, we will assign the sentences to each of these detected (or inferred, in the "(A-D)" case) sub-figure identifiers. The \textit{opening common} is usually well-preserved, while the \textit{closing common} may inevitably contain some information dedicated to the last mentioned identifier. In order to eliminate the risk of potential information loss caused by excluding the \textit{closing common} texts, we compromise to this heuristic to include the last few sentences.

\paragraph{Preprocessing}
We also conduct preprocessing and text cleansing during the sub-caption segmentation. The extracted word blocks sometimes contain \textit{dashed} words where the dash is inserted among a single word due to the line change in PDF articles. We replace these dashes by checking if the word with the dash removed exists in a common English dictionary. For those words that do not exist, we simply preserve the dash within the words. We also run spell checkers to ensure spellings for common English words are correct as much as possible.

\subsection{More Exemplar Data Points}
\label{a-ssec: more_exps}
\begin{figure}[ht!]

\begin{subtable}{.47\columnwidth}
{\tiny \textbf{\texttt{Experiment Method Labels}}}

\vspace{1mm}

\fontsize{2mm}{5} \selectfont \texttt{Par(coarse)\hspace{1mm}:\hspace{0.5mm}unassigned}

\fontsize{2mm}{5} \selectfont \texttt{\hspace{15mm} \space}

\vspace{1mm}

\fontsize{2mm}{5} \selectfont \texttt{Par(fine)\hspace{3.5mm}:\hspace{0.5mm}experimental}

\fontsize{2mm}{5} \selectfont \texttt{\hspace{15mm} participant}

\vspace{1mm}

\fontsize{2mm}{5} \selectfont \texttt{Int(coarse)\hspace{1mm}:\hspace{0.5mm}biophysical}

\fontsize{2mm}{5} \selectfont \texttt{\hspace{15mm} \space}

\vspace{1mm}

\fontsize{2mm}{5} \selectfont \texttt{Int(fine)\hspace{3.5mm}:\hspace{0.5mm}molecular}

\fontsize{2mm}{5} \selectfont \texttt{\hspace{15mm} sieving}

\vspace{2mm}

\centering
  \includegraphics[width=\columnwidth]{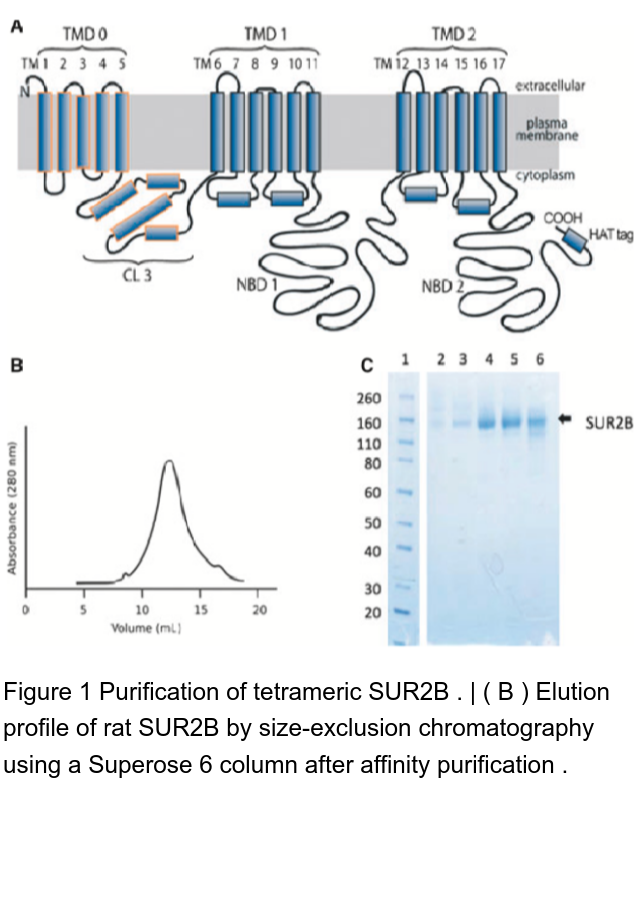}
\end{subtable}%
\quad
\begin{subtable}{.47\columnwidth}
{\tiny \textbf{\texttt{Experiment Method Labels}}}

\vspace{1mm}

\fontsize{2mm}{5} \selectfont \texttt{Par(coarse)\hspace{1mm}:\hspace{0.5mm}identification}

\fontsize{2mm}{5} \selectfont \texttt{\hspace{15mm} by antibody}

\vspace{1mm}

\fontsize{2mm}{5} \selectfont \texttt{Par(fine)\hspace{3.5mm}:\hspace{0.5mm}anti-tag}

\fontsize{2mm}{5} \selectfont \texttt{\hspace{15mm} western}

\vspace{1mm}

\fontsize{2mm}{5} \selectfont \texttt{Int(coarse)\hspace{1mm}:\hspace{0.5mm}affinity}

\fontsize{2mm}{5} \selectfont \texttt{\hspace{14.5mm} chromatography}

\vspace{1mm}

\fontsize{2mm}{5} \selectfont \texttt{Int(fine)\hspace{3.5mm}:\hspace{0.5mm}anti-tag}

\fontsize{2mm}{5} \selectfont \texttt{\hspace{15mm} coip}

\vspace{2mm}

\centering
  \includegraphics[width=\columnwidth]{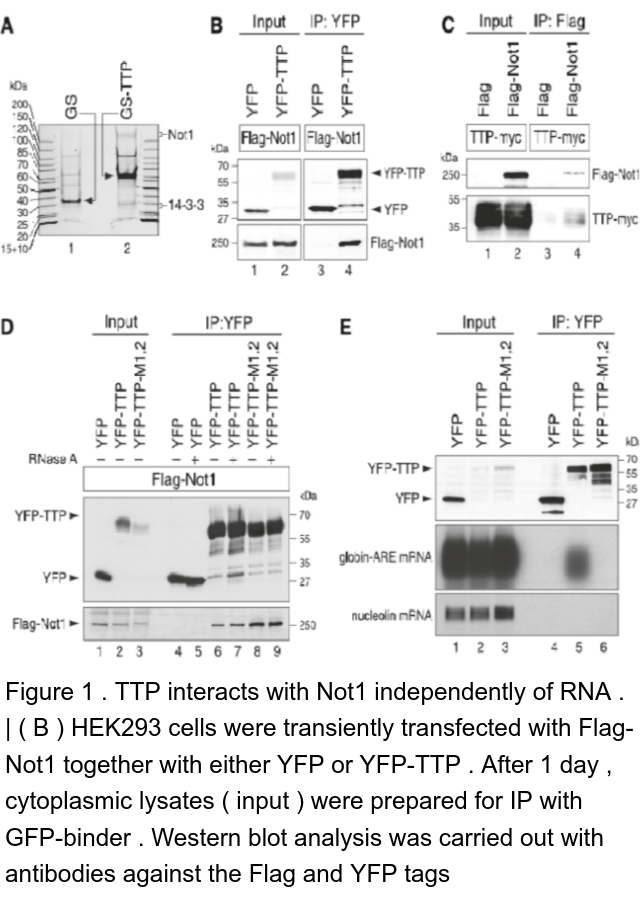}
 
\end{subtable}


\begin{subtable}{.47\columnwidth}
\vspace{3mm}

{\tiny \textbf{\texttt{Experiment Method Labels}}}

\vspace{1mm}

\fontsize{2mm}{5} \selectfont \texttt{Par(coarse)\hspace{1mm}:\hspace{0.5mm}nucleotide}

\fontsize{2mm}{5} \selectfont \texttt{\hspace{15mm} sequence}

\vspace{1mm}

\fontsize{2mm}{5} \selectfont \texttt{Par(fine)\hspace{3.5mm}:\hspace{0.5mm}southern}

\fontsize{2mm}{5} \selectfont \texttt{\hspace{15mm} blot}

\vspace{1mm}

\fontsize{2mm}{5} \selectfont \texttt{Int(coarse)\hspace{1mm}:\hspace{0.5mm}cross-linking}

\fontsize{2mm}{5} \selectfont \texttt{\hspace{15mm} study}

\vspace{1mm}

\fontsize{2mm}{5} \selectfont \texttt{Int(fine)\hspace{3.5mm}:\hspace{0.5mm}cross-link}

\fontsize{2mm}{5} \selectfont \texttt{\hspace{15mm} \space}

\vspace{2mm}

\centering
  \includegraphics[width=\columnwidth]{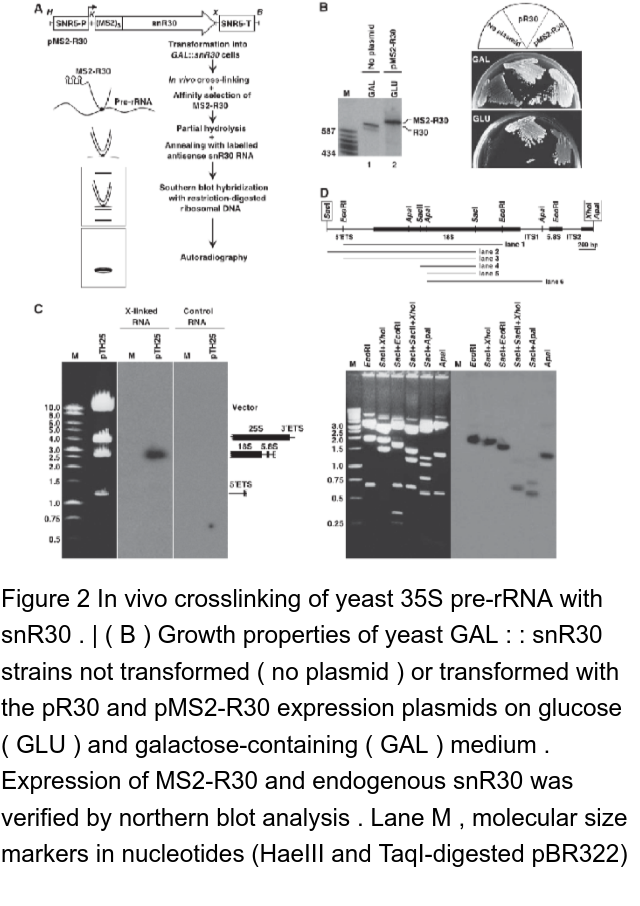}
\end{subtable}%
\quad
\begin{subtable}{.47\columnwidth}
\vspace{3mm}

{\tiny \textbf{\texttt{Experiment Method Labels}}}

\vspace{1mm}

\fontsize{2mm}{5} \selectfont \texttt{Par(coarse)\hspace{1mm}:\hspace{0.5mm}identification}

\fontsize{2mm}{5} \selectfont \texttt{\hspace{15mm} by antibody}

\vspace{1mm}

\fontsize{2mm}{5} \selectfont \texttt{Par(fine)\hspace{3.5mm}:\hspace{0.5mm}anti-tag}

\fontsize{2mm}{5} \selectfont \texttt{\hspace{15mm} western}

\vspace{1mm}

\fontsize{2mm}{5} \selectfont \texttt{Int(coarse)\hspace{1mm}:\hspace{0.5mm}affinity}

\fontsize{2mm}{5} \selectfont \texttt{\hspace{14.5mm} chromatography}

\vspace{1mm}

\fontsize{2mm}{5} \selectfont \texttt{Int(fine)\hspace{3.5mm}:\hspace{0.5mm}anti-tag}

\fontsize{2mm}{5} \selectfont \texttt{\hspace{15mm} coip}

\vspace{2mm}

\centering
  \includegraphics[width=\columnwidth]{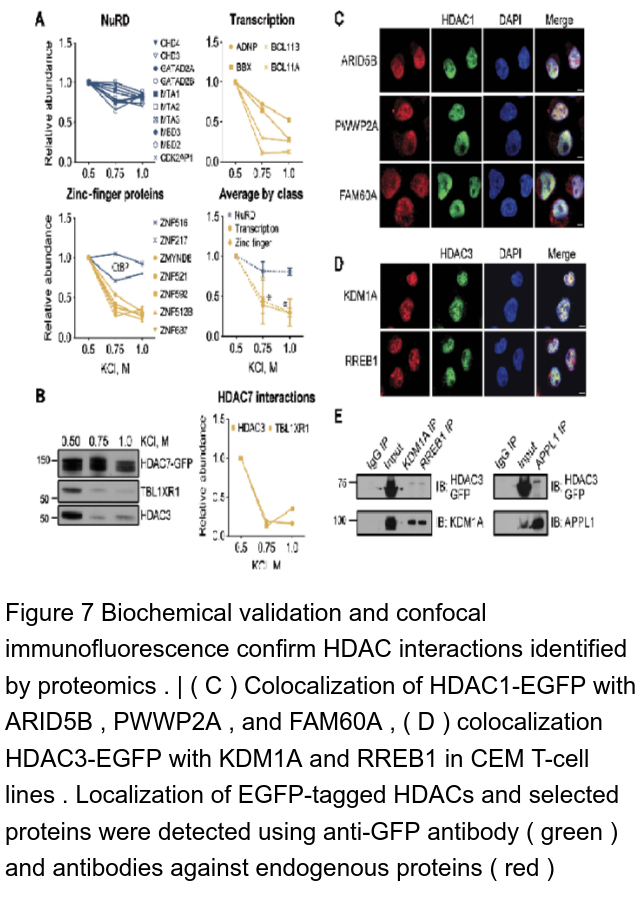}
 
\end{subtable}

\centering
    \caption{
        \textbf{More sample data:} descriptions of the data structure of \dataset~dataset can be referred to in~\figref{fig:data_samples}.
    }
    \label{fig:appendix_data_samples}
\end{figure}
\noindent We provide more random sampled exemplar data for a more diverse visual preview of our dataset in~\figref{fig:appendix_data_samples}.

\vspace{0.3em}

\mypar{Expert Justifications of Requirements of Multimodality:} As discussed in the third paragraph of the introduction in~\secref{sec:intro}, different experiment methods can generate visually similar figures, while captions can disambiguate them. Oppositely, captions alone do not necessarily determine the visual presentation to be of specific types. Specifically, for the IntAct database used as distant supervision to construct our dataset, in the IntAct curation annotation manual\footnote{\url{https://www.ebi.ac.uk/~intact/site/doc/IntActCurationManual2016.pdf?conversationContext=1}} page 5, the storage schema implies at the experiment level curators are \textbf{required} to look at the paper figures. And in page 9, the example term \textit{"From Fig 7..."} indicates that curators are expected to interpret both the figures and captions for labelling. There are similar examples throughout the manual, which are verified by an expert in the biomedical field.

\vspace{0.3em}

\mypar{Sample Human Estimations:} There are various cues that a human expert would likely to focus on when determining the experiment method labels. For example, for the label \textit{identification-by-antibody}, the presence of specific bands in a blot diagram corresponding to the precise molecular weight of the named antibody in the captions, would be the key features to look for, as hinted in~\figref{fig:appendix_data_samples} right sides.
We also hereby provide an expert point of view of how to infer the experiment method labels from the given multimodal inputs. In~\figref{fig:appendix_data_samples} bottom right, the caption says \textit{"(C) Colocalization of HDAC1-EGFP..."} and the figure clearly shows fluorescent images. The statement \textit{colocalization} should, but does not necessarily determine that figures will be of specific types. The presentation of cell-level \textit{colabeling} images provides enormously more context than that found in the caption. Thus the visual appearance of figures is an important signal for scientists, and the captions usually provide abbreviated crucial descriptions of the methods used.

\subsection{Label Distributions}

As hinted by~\figref{fig:label_dist}, the label distribution in the original IntAct records is imbalanced, which very likely resemble the real world distribution of the experimental protocols being conducted. This could pose challenges and inspire interesting future directions in machine learning models on tackling these domain specific areas. Our non-processed raw data has the label distribution on the \textit{Par(Coarse)} label type shown in~\tbref{tab:label_dist} (The shown numbers are of label-count percentages).
We would also like to point out that the finer-grained classes, such as \textit{Int(Fine)} have more even distribution. Since there are in total 85 classes for \textit{Int(Fine)}, we hereby show the mean and the standard deviations of the label-count percentages, which are: mean/std = 1.18/3.50 as compared to 14.29/21.18 of 7 classes in \textit{Par(Coarse)}.
On the other hand, we also show in~\tbref{tab:mod-comp} the performances of the majority baselines (always predicting the majority class) are significantly worse than the best performing models, which implies proper training could potentially still yield good performances on the current data distributions. We do hope to expand our dataset to an even larger-scaled and possibly more balanced one when more publicly available articles are added to the OA-PMC set, particularly for the labels that are on the lower amount spectrum.

\begin{table}[t]
\centering
\footnotesize
\begin{tabular}{r|c}
    \toprule
    \multicolumn{1}{c}{\textbf{Label}} & \textbf{Count (\%)} \\
    \midrule
    predetermined-participant & 58.96 \\
    identification-by-antibody & 33.63 \\
    nucleotide-sequence-identification & 4.56 \\
    Identification-by-mass-spectrometry & 1.02 \\
    unassigned & 0.61 \\
    tag-visualisation & 0.94 \\
    protein-sequence-identification & 0.28 \\
    \bottomrule
\end{tabular}
\caption{\textbf{Label Distributions:} of the raw IntAct records.}
\label{tab:label_dist}
\end{table}

\section{More Details on Experiments}
\label{a-sec: more_exps}

\subsection{Model \& Training Details}
\label{a-ssec: more_details}

All the benchmarked models are trained on a single Nvidia GeForce 2080Ti GPU\footnote{https://www.nvidia.com/en-us/geforce/graphics-cards/rtx-2080-ti/} on a CentOS 7 operating system.
The hyper-parameters for each model are manually tuned against our dataset, and the trained model checkpoints used to evaluate are selected by the best performing ones on the validation set.
All the models are individually trained for each type of the labels.

The implementations of the transformer-based caption-only models are extended from the huggingface\footnote{\url{https://github.com/huggingface/transformers}}~code base, which are implemented in PyTorch\footnote{https://pytorch.org/}.
The image-only models and their pretrained weights are borrowed from torchvision\footnote{https://github.com/pytorch/vision} and the official Mask-RCNN implementations from Facebook AI Research\footnote{https://github.com/facebookresearch/detectron2}. Note that we are just using the pretrained weights to initialize only the CNN part, \ie we are not using the ROI pooled features, so in our image-only models, \textbf{no region proposal network (RPN)} is used. Implementations for both VL-BERT and ViL-BERT are adapted from the original author-released code repositories, which can be found in their papers.

\paragraph{Visual-MLM}
For the masked image ROI region classification or masked visual token modeling in VL-BERT and ViL-BERT, dubbed as visual-MLM in this paper, we use the authors' original public repositories for obtaining the region labels in the proposed ROIs in the images. Since the two repositories both generally adopt \textit{detectron2} module from Facebook AI Research, the label space comes from the Visual Genome~\cite{krishna2017visual} object categories, which has 1,600 total number of class labels, plus one indicating the background class.


\subsubsection{Hyper-parameters}
\label{a-ssec:hyparams}

\begin{table*}[t]
\centering
\scriptsize
    \begin{tabular}{clccccccc}
    \toprule
    \multirow{2}{*}{\textbf{Modalities}} & \multicolumn{1}{c}{\multirow{2}{*}{\textbf{Models}}} & \multirow{2}{*}{\textbf{Batch Size}} & \multirow{2}{*}{\textbf{Initial LR}} & \multirow{2}{*}{\textbf{\# Training Epochs}} & \textbf{Gradient Accu-} & \multirow{2}{*}{\textbf{\# Params}}  \\
    & & & & & \textbf{mulation Steps} & \\
    \midrule
    \multirow{2}{*}{Image-Only}
    & ResNet101 init. from ImageNet & 16 & $1 \times 10^{-5}$ & 20 & - & 44.5M \\
    & ResNet101 init. from MSCoCo & 16 & $1 \times 10^{-5}$ & 20 & -  & 44.5M \\
    \hline \\[-1em]
    \multirow{3}{*}{Caption-Only}
    & LSTM w. BioGlove & 4 & $2 \times 10^{-6}$ & 4  & 1 & 5.4M  \\
    & RoBERTa-Large & 4 & $2 \times 10^{-6}$ & 4  & 2 & 355.4M \\
    & SciBERT-Base-Uncased & 4 & $2 \times 10^{-6}$ & 4 & 2 & 109.9M \\
    \hline \\[-1em]
    \multirow{7}{*}{Multi-Modal}
    & Naive Late Fusion (NLF) & 4 & $2 \times 10^{-6}$ & 4  & 2 & 126.5M \\
    & Stacked Attention Network (SAN) & 4 & $2 \times 10^{-6}$ & 4  & 2 & 130.8M \\
    & ViL-BERT (w. MLM) & 8 & $1 \times 10^{-5}$ & 20  & 1 & 171.2M \\
    & ViL-BERT (w. MLM \& NSP) & 8 & $1 \times 10^{-5}$ & 20  & 1 & 171.2M \\
    & ViL-BERT ((w. MLM \& NSP \& visual-MLM) & 8 & $1 \times 10^{-5}$ & 20  & 1 & 171.2M \\
    & VL-BERT (w. MLM) & 4 & $7 \times 10^{-5}$ & 20  & 4 & 155.6M \\
    & VL-BERT (w. MLM \& visual-MLM) & 2 & $7 \times 10^{-5}$ & 20  & 4 & 155.6M \\
    \bottomrule
    \end{tabular}
\caption{\textbf{Hyper-parameters used for each of our baseline models during finetuning phase on our \dataset{} dataset:} For each label type (coarse and fine \textit{participant} and \textit{interaction}), we use the same set of hyperparameters. \textit{Initial LR} denotes initial learning rate. All the models are trained with Adam optimizers
. We include number of prarameters of each model in the last column, denoted as \textit{\# params}.}
\label{tab:hyparams}
\end{table*}

\begin{table*}[t]
\centering
\scriptsize
    \begin{tabular}{clcccccc}
    \toprule
    \multirow{2}{*}{\textbf{Modalities}} & \multicolumn{1}{c}{\multirow{2}{*}{\textbf{Models}}} & \multirow{2}{*}{\textbf{Batch Size}} & \multirow{2}{*}{\textbf{Initial LR}} & \multirow{2}{*}{\textbf{\# Training Epochs}} & \textbf{Gradient Accu-} \\
    & & & & & \textbf{mulation Steps} \\
    \midrule
    \multirow{2}{*}{Caption-Only}
    & RoBERTa-Large & 4 & $1 \times 10^{-5}$ & 50  & 1 \\
    & SciBERT-Base-Uncased & 4 & $1 \times 10^{-5}$ & 50  & 1 \\
    \hline \\[-1em]
    \multirow{5}{*}{Multi-Modal}
    & ViL-BERT (w. MLM) & 8 & $5 \times 10^{-6}$ & 20  & 1  \\
    & ViL-BERT (w. MLM \& NSP) & 8 & $5 \times 10^{-6}$ & 20  & 1  \\
    & ViL-BERT (w. MLM \& NSP \& visual-MLM) & 8 & $5 \times 10^{-6}$ & 20  & 1  \\
    & VL-BERT (w. MLM) & 4 & $7 \times 10^{-5}$ & 20  & 4   \\
    & VL-BERT (w. MLM \& visual-MLM) & 2 & $7 \times 10^{-5}$ & 20  & 4  \\
    \bottomrule
    \end{tabular}
\caption{\textbf{Hyper-parameters used for language models with~\dataset~corpus finetuning phase:} the hyperparameters for caption-only models are also adopted for the two non-transformer based multimodal models for their language part encoders. All the models are trained with Adam optimizers.}
\label{tab:hyparams-pre}
\end{table*}

\begin{table*}[t]
\centering
\footnotesize
\begin{tabular}{ccccc}
    \toprule
    \textbf{Type} & \textbf{Batch Size} & \textbf{Initial LR} & \textbf{\# Training Epochs} & \textbf{Gradient Accumulation Steps} \\
    \midrule
    \textbf{Bound (lower--upper)} & 2--16 & $1 \times 10^{-4}$--$1 \times 10^{-6}$ & 3--50 & 1--4 \\
    \midrule
    \textbf{Number of Trials} & 2--4 & 2--3 & 2--4 & 1--2 \\
    \bottomrule
\end{tabular}
\caption{\textbf{Search bounds:} for the hyperparameters of all the benchmarked models.}
\label{tab:search}
\end{table*}

For image-only models and caption-only models, it takes roughly 2-4 hours to train for the number of epochs specified in~\tbref{tab:hyparams}. For NLF and SAN models, it takes approximately 4 hours to train. For the two visual-linguistics models, it takes roughly 6-8 hours to train including the finetuning on experiment classification phases. Since some of the baseline models incorporate finetuning on our~\dataset~corpus, we also list the hyperparameters such phase of training uses in~\tbref{tab:hyparams-pre}. We also include the search bounds and number of trials in~\tbref{tab:search}, all of our models adopt the same search bounds and the same ranges of trials.

For long captions in the dataset, we generally truncate it with maximum number of tokens (BPE or wordpiece tokens) $=128$.

\subsubsection{Vision Models}
As mentioned in~\secref{sec:baselines}, the \textit{benchmark models} section, we only finetune the later layers in ResNet-101, due to the observations that the features in early CNN layers are generally shared across different visual domains~\cite{yosinski2014transferable}. We also empirically verify that finetuning all the layers does not yield significant improvements.

\subsubsection{Validation Results}
\label{a-ssec:dev_res}

We also include the model performances on the validation set, where we select each of the best performing models to evaluate on the unseen test set. The comprehensive results can be found in~\tbref{tab:mod-comp-dev}.

\begin{table*}[t!]
\centering
\small
    \begin{tabular}{cclcccccc}
    \toprule
    \textbf{Modalities} & \multicolumn{1}{c}{\textbf{Models}} & \multicolumn{1}{c}{\textbf{Variants}} & \textbf{Par$_{\textbf{coarse}}$} & \textbf{Int$_{\textbf{coarse}}$} & \textbf{Par$_{\textbf{fine}}$} & \textbf{Int$_{\textbf{fine}}$} \\
    \midrule \\[-1.1em]
    
    \multicolumn{1}{c}{---}
    & Majority Baseline
    & \multicolumn{1}{c}{---} & 62.36 & 63.02 & 57.68  & 22.04  \\

    \midrule \\[-1.1em]    
    \multirow{2}{*}{Image-Only}
    
    & \multirow{2}{*}{ResNet-101}
    & init. from ImageNet & 68.52 & 70.98 & 58.56  & 24.07  \\
    & & init. from MSCoCo & 64.73 & 69.42 & 56.02  & 26.15  \\
    \midrule \\[-1.1em]
    
    \multirow{5}{*}{Caption-Only}
    
    & LSTM w. BioGloVe & \multicolumn{1}{c}{---} & 64.37 & 66.82 & 58.35  & 33.41  \\
    \cline{2-7} \\[-.8em]

    & \multirow{2}{*}{RoBERTa}
    & w/o MLM finetuning  & \textbf{78.20} & 81.50 & 65.70  & 60.36  \\ 
    & & w. MLM finetuning & 77.73 & 83.74 & 67.04  & 61.47 \\
    \cline{2-7} \\[-.8em]

    & \multirow{2}{*}{SciBERT}
    & w/o MLM finetuning & 77.30 & 86.40 & 62.10  & 65.70  \\ 
    & & w. MLM finetuning & 74.17 & 86.19 & 65.70  & 58.13 \\ 
    \midrule \\[-1.1em]
    
    \multirow{9}{*}{Multi-Modal}

    & \multirow{2}{*}{NLF}
    & w/o language part MLM finetuning & 75.95 & 86.90 & 67.04  & 58.80  \\ 
    & & w. language part MLM finetuning & 75.49 & 87.08 & 65.92  & 61.02 \\
    \cline{2-7} \\[-.8em]

    & \multirow{2}{*}{SAN}
    & w/o language part MLM finetuning & 77.70 & 86.90 & \textbf{67.48}  & 59.69  \\ 
    & &  w. language part MLM finetuning & 74.60 & 87.50 & 65.92  & 59.47 \\ 
    \cline{2-7} \\[-.8em]
    
    & \multirow{3}{*}{ViL-BERT}
    & w. MLM & 72.83 & 85.97 & 65.92  & 63.25  \\
    & & w. MLM \& NSP & 75.28 & 86.86 & 62.36  & 63.03  \\
    & & w. MLM \& NSP \& visual-MLM & 73.05 & 87.31 & 61.69  & 63.03  \\
    \cline{2-7} \\[-.8em]
    
    & \multirow{2}{*}{VL-BERT}
    & w. MLM & 76.05 & \textbf{87.52} & 66.91  & \textbf{67.48}  \\
    & & w. MLM \& visual-MLM & 75.49 & 87.19 & 65.52 & 66.86 \\
    
    \bottomrule
    \end{tabular}
\caption{\textbf{Model accuracies on the validation set:} the two label categories are denoted as \textit{Par}, and \textit{Int} for \textit{participant} and \textit{interaction} method respectively. The label hierarchy is indicated as the subscript, \eg Par$_{\textbf{coarse}}$ indicates coarse types of \textit{participant} method. The best performances for each type of label are bolded. In most cases, multimodal models outperform the unimodal models except for the coarse \textit{participant} type.}
\label{tab:mod-comp-dev}
\end{table*}

\subsection{More Attention Visualizations}
\label{a-ssec:multi_model_attn}
\begin{figure*}[ht]
\footnotesize
\begin{tabular}{cccc}

\includegraphics[width=.47\columnwidth]{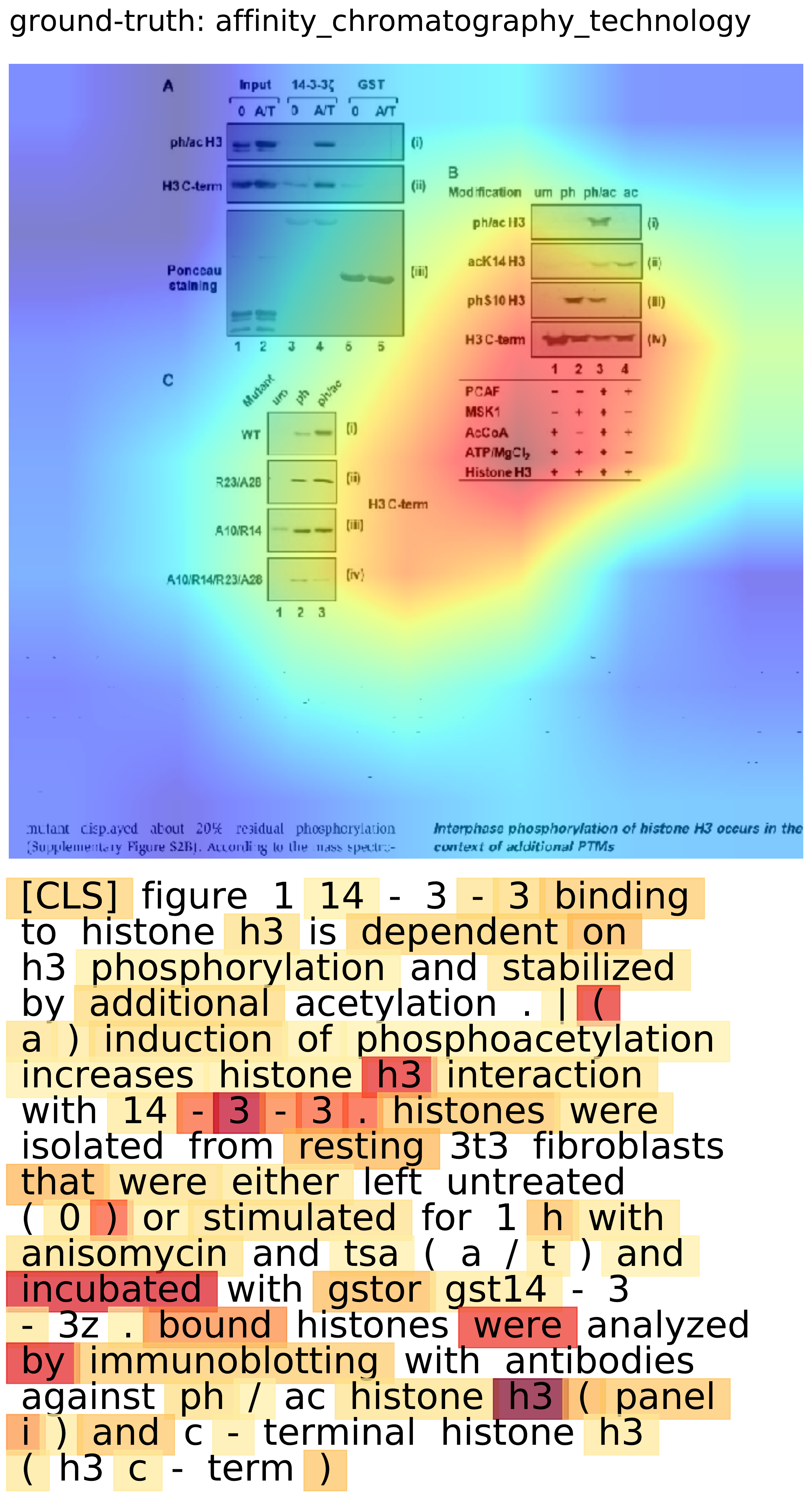}
& \includegraphics[width=.47\columnwidth]{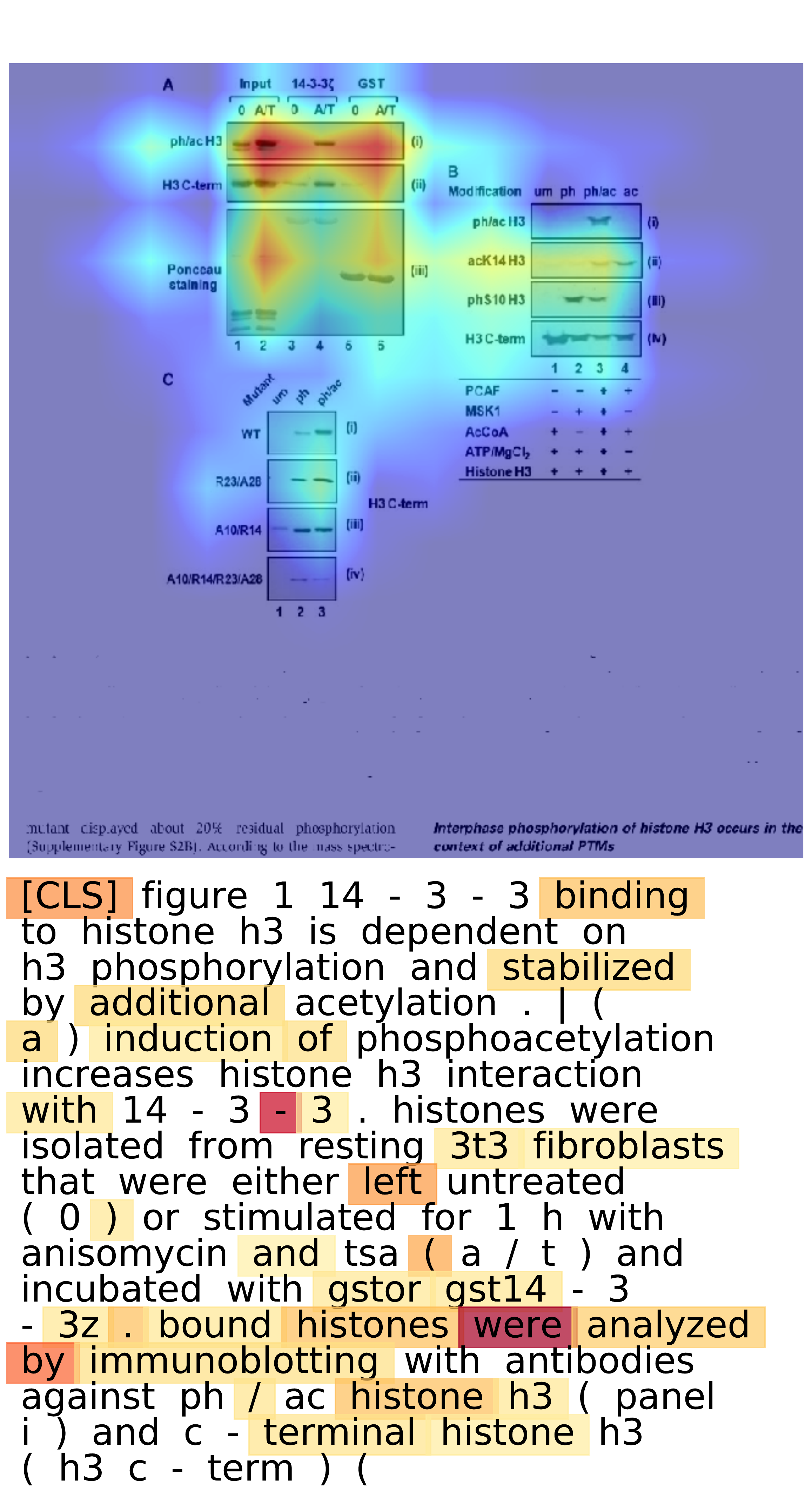} 
& \includegraphics[width=.47\columnwidth]{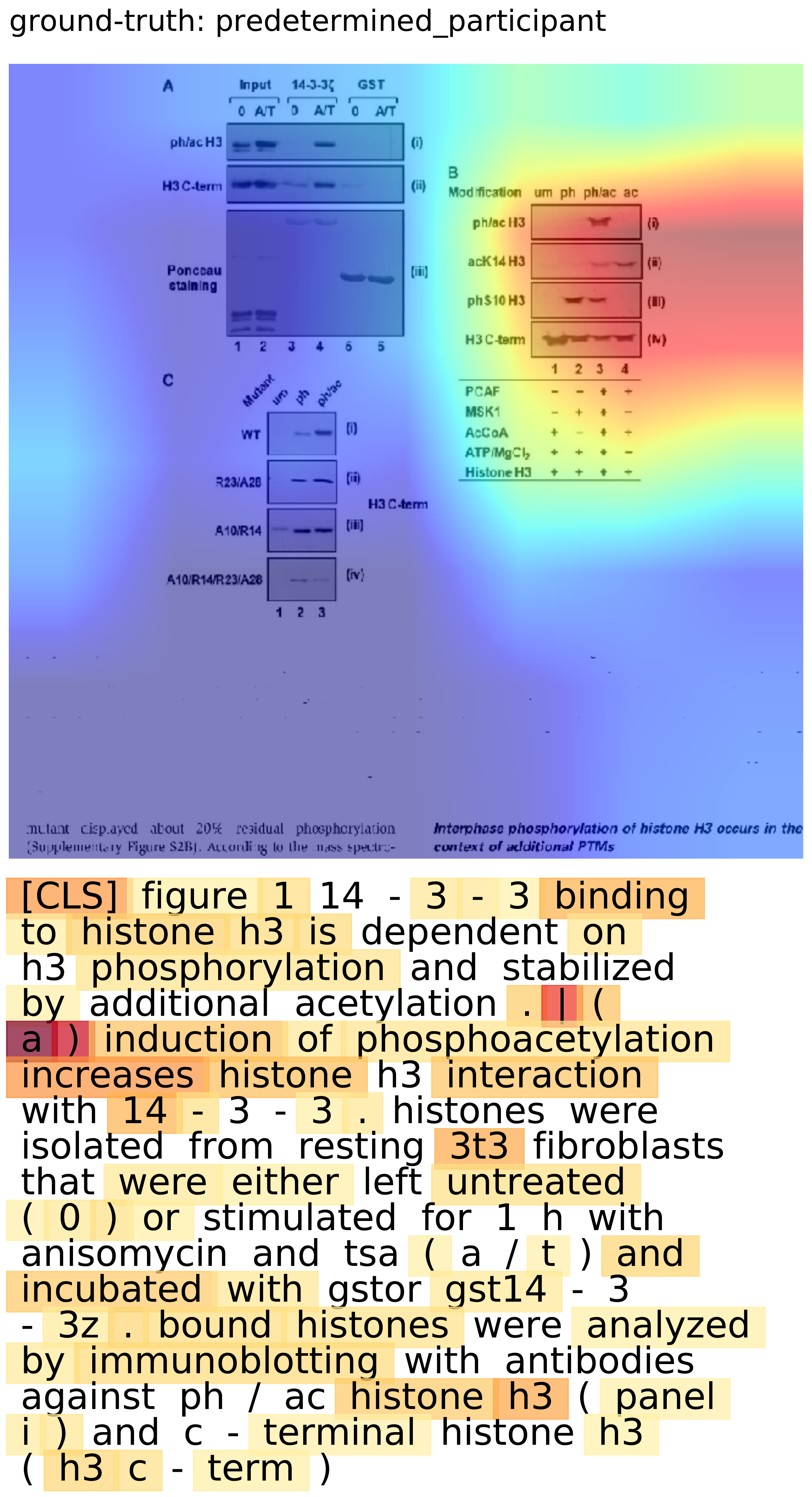}
& \includegraphics[width=.47\columnwidth]{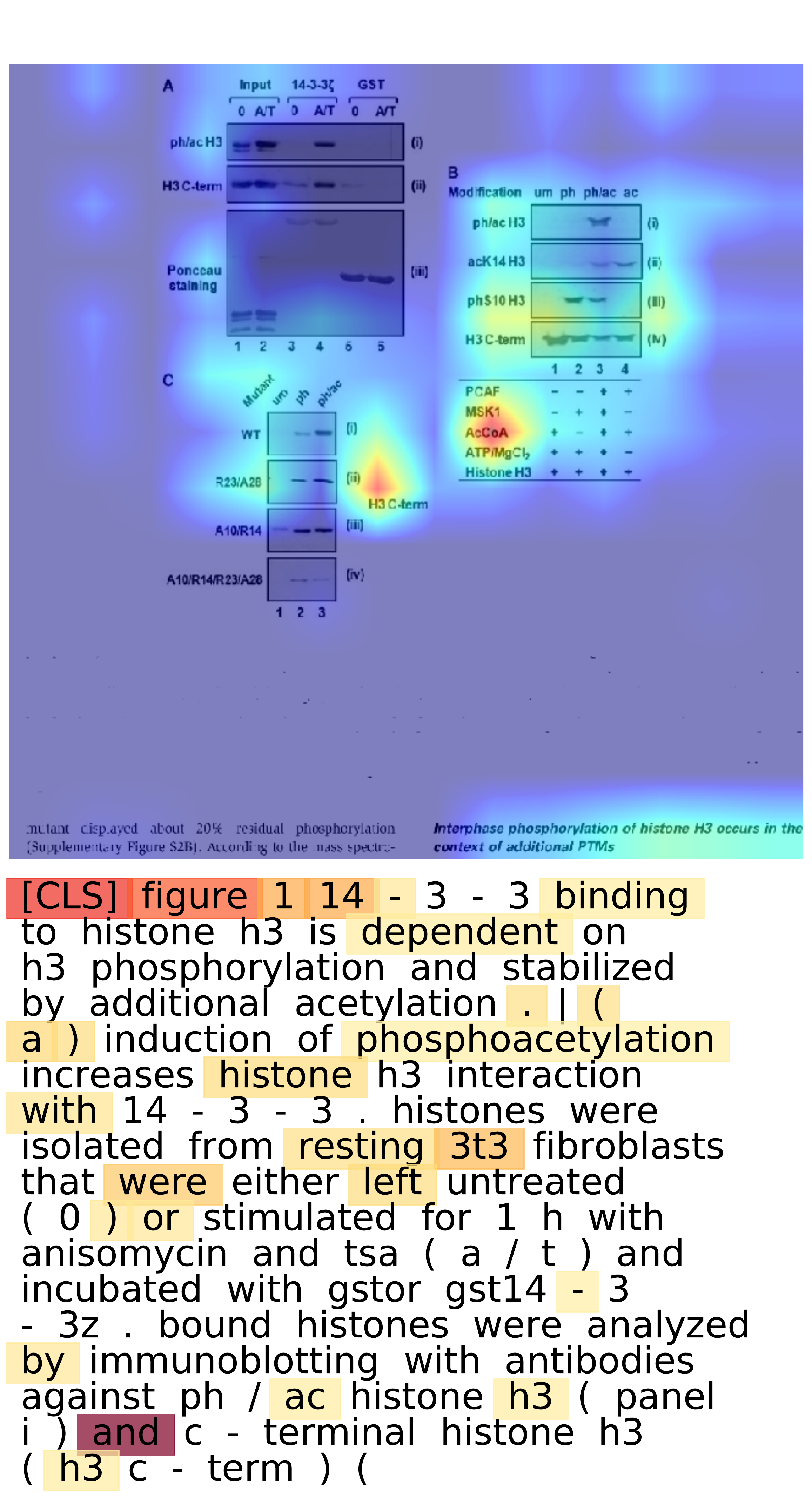} \\
\\ [-.5em]

\includegraphics[width=.47\columnwidth]{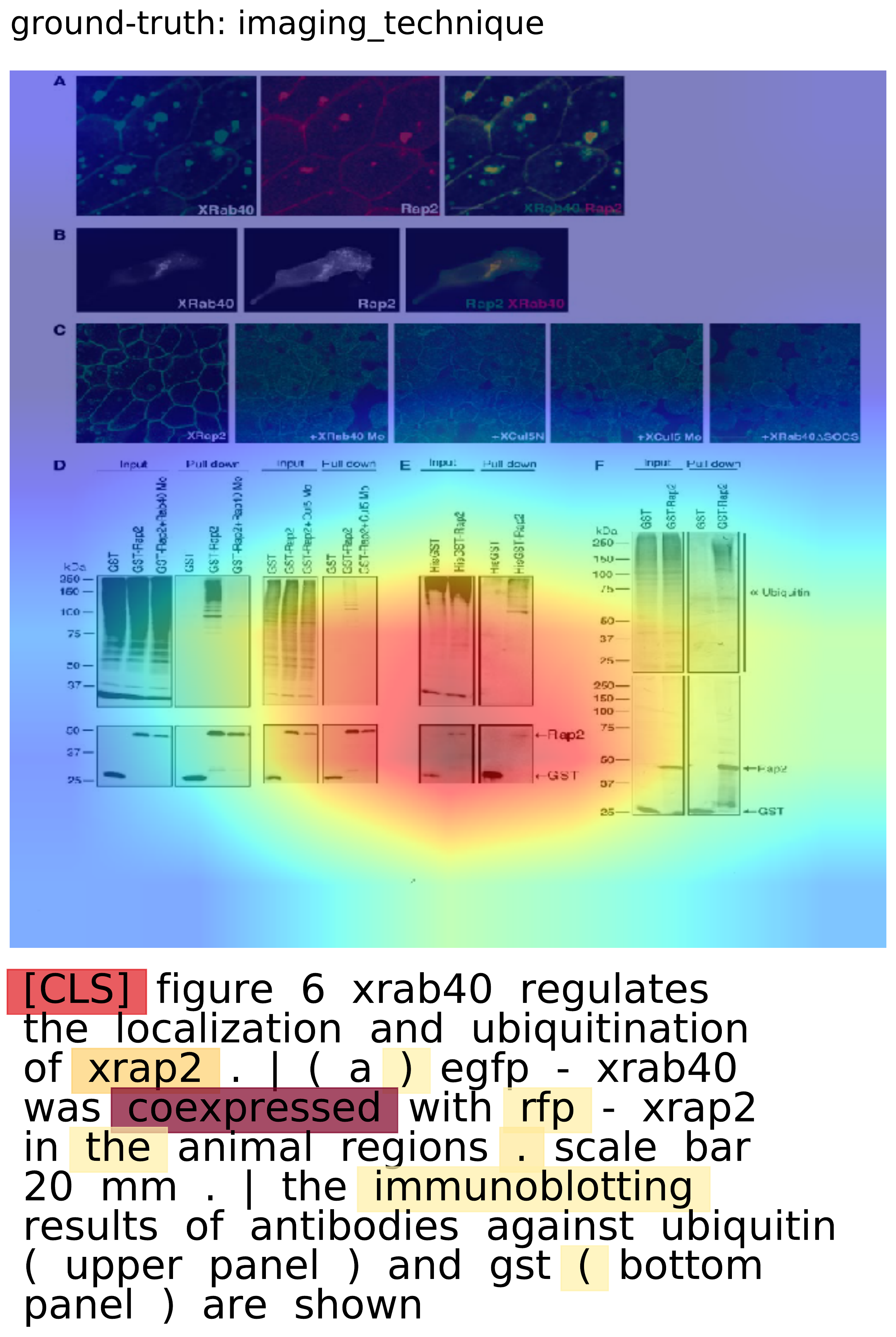} 
& \includegraphics[width=.47\columnwidth]{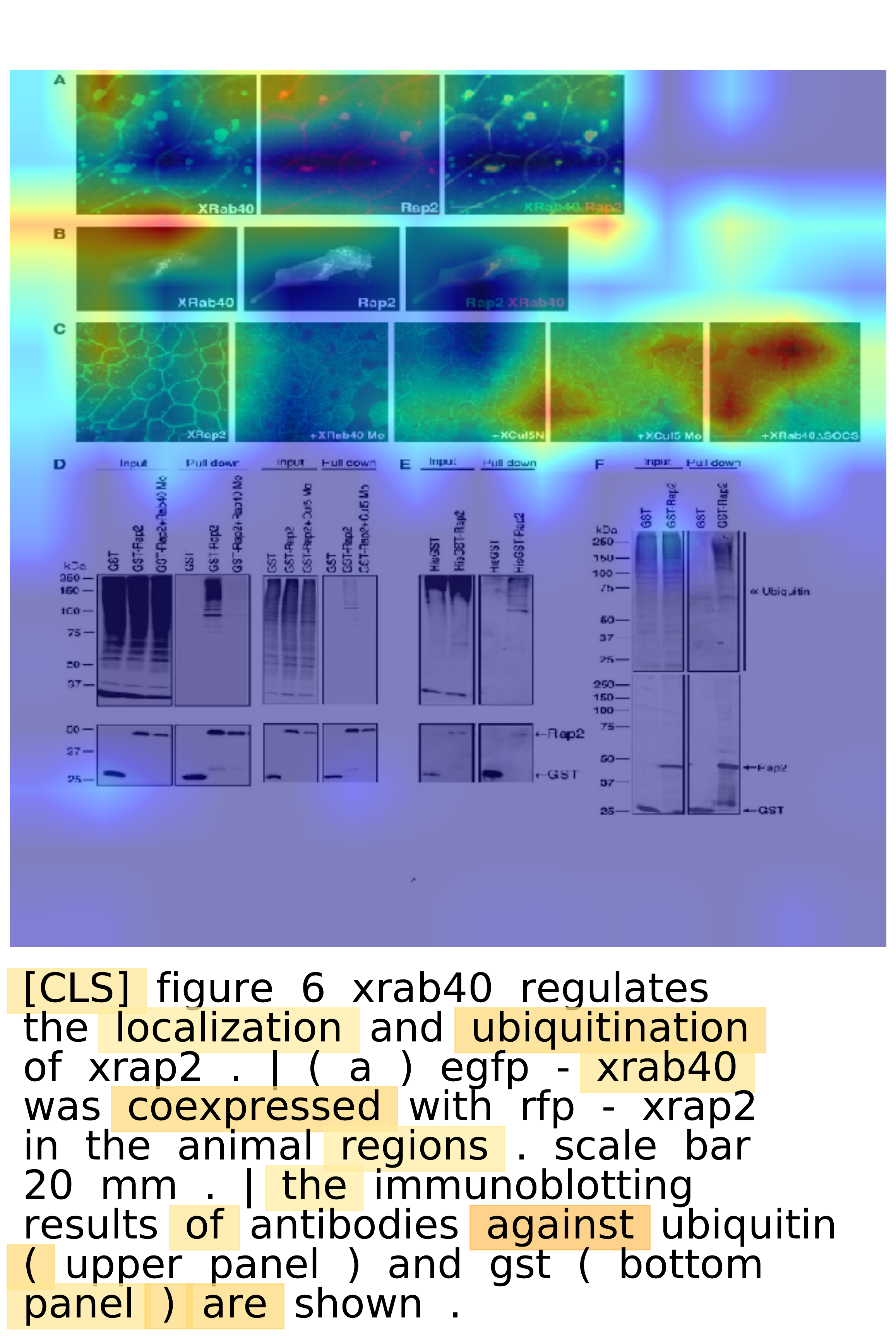}
& \includegraphics[width=.47\columnwidth]{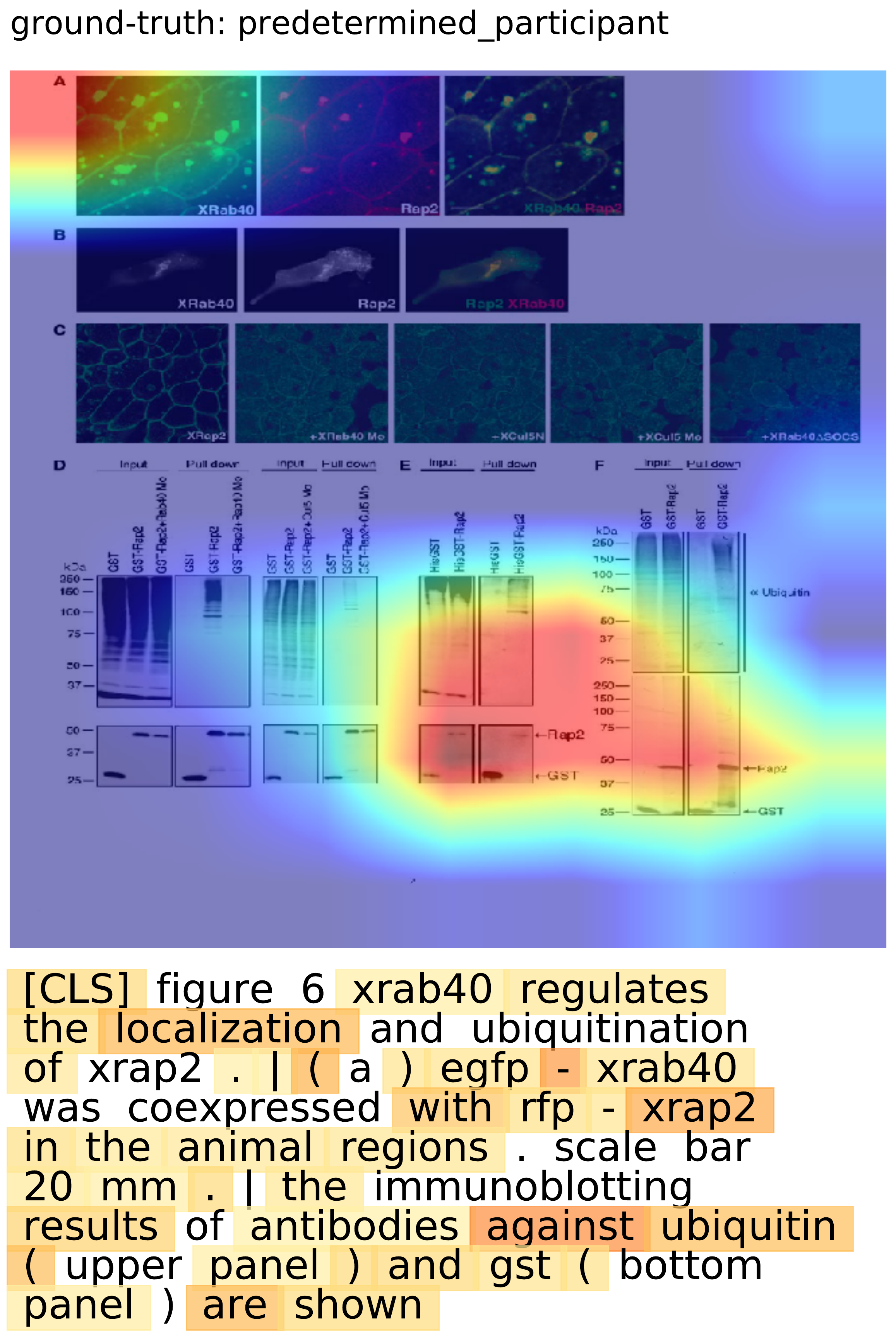}
& \includegraphics[width=.47\columnwidth]{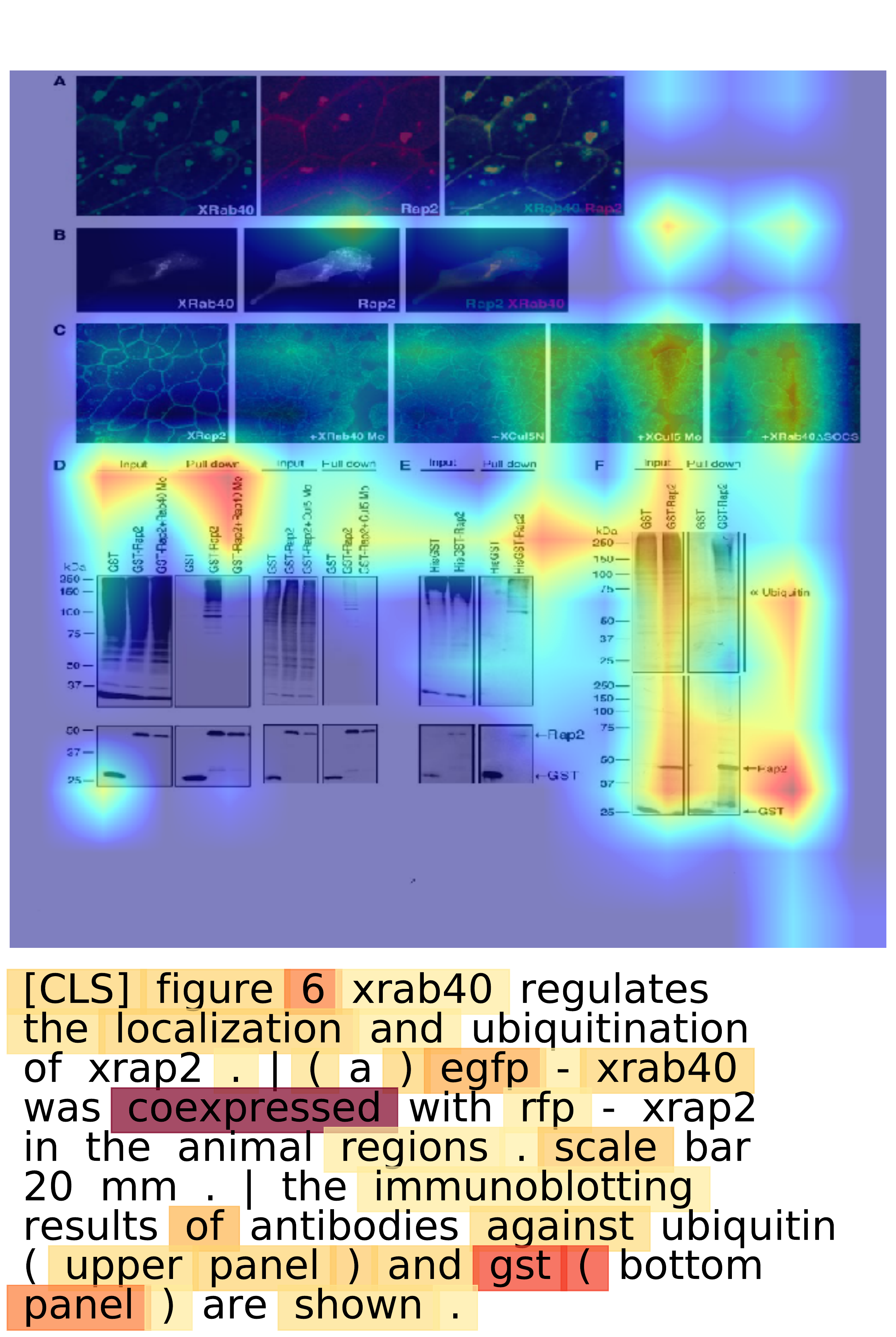} \\
\\ [-.5em]

\includegraphics[width=.47\columnwidth]{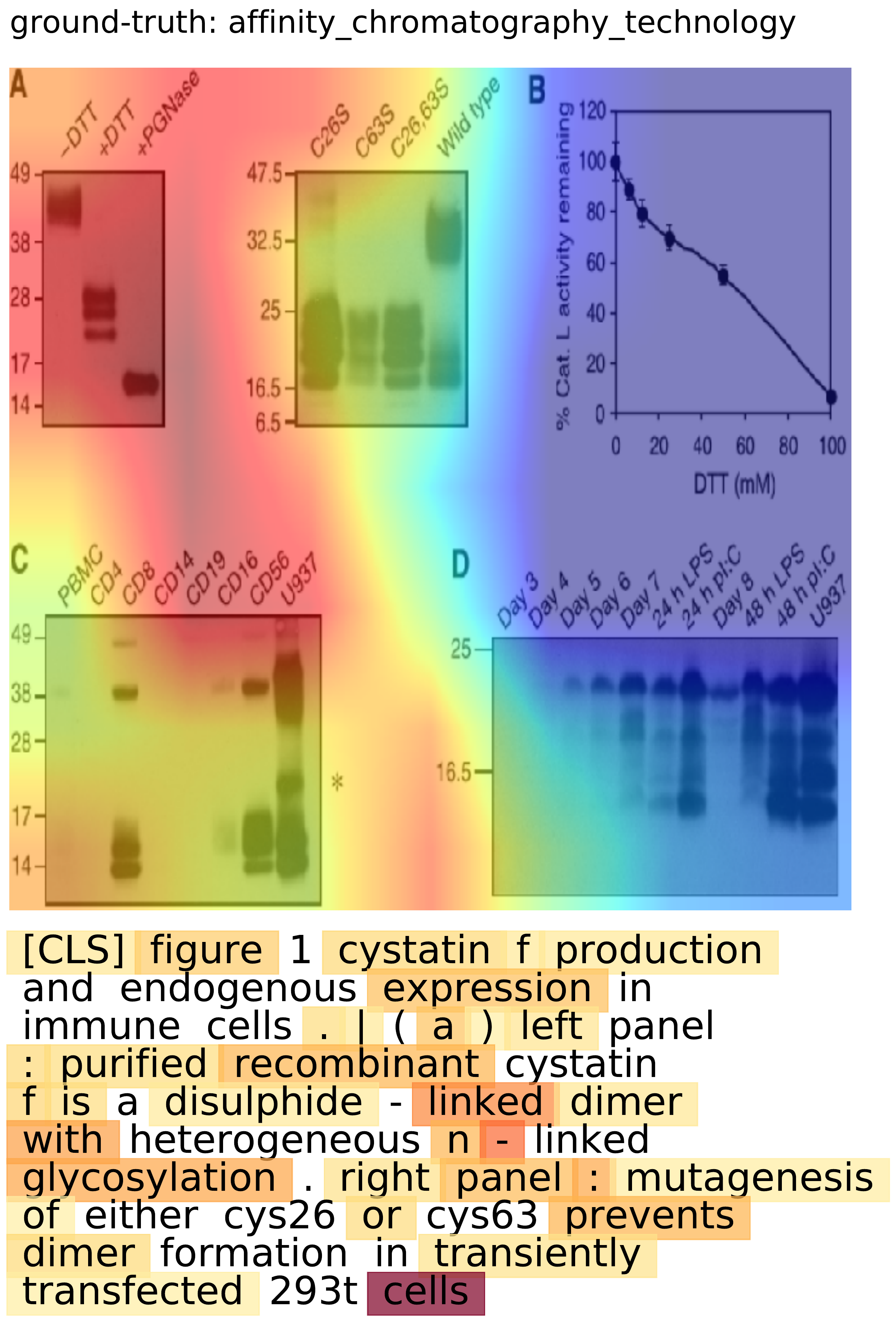} 
& \includegraphics[width=.47\columnwidth]{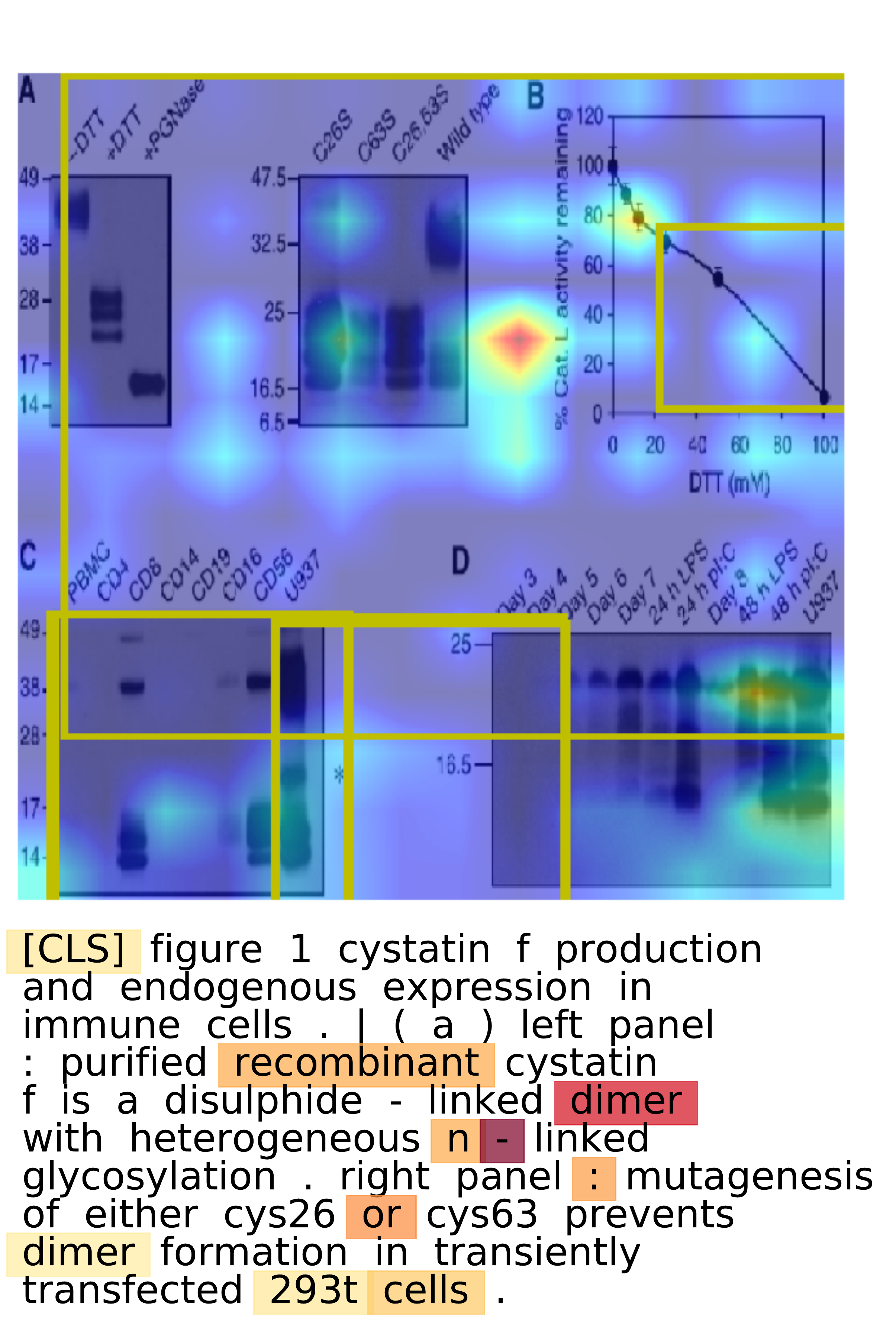}
& \includegraphics[width=.47\columnwidth]{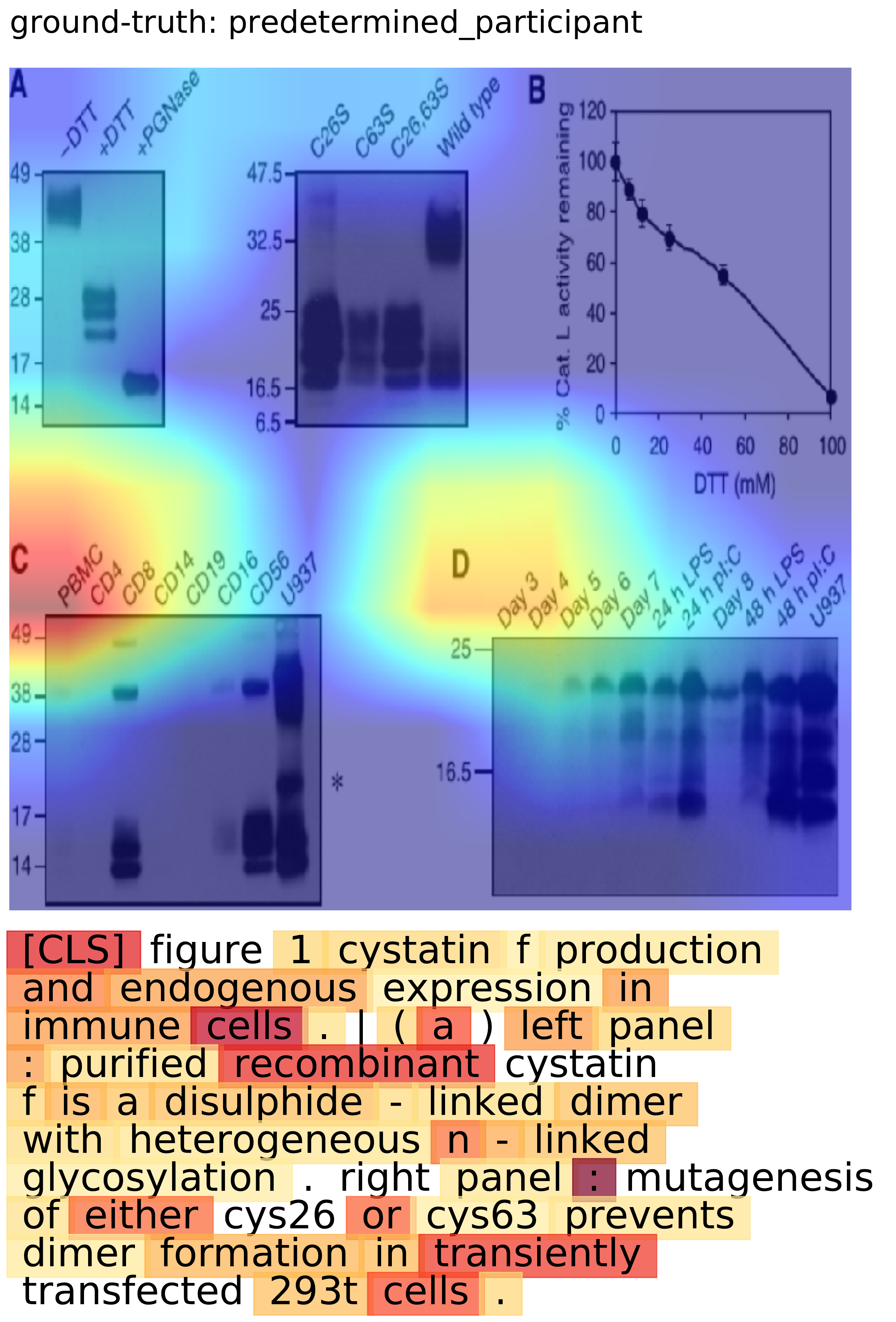}
& \includegraphics[width=.47\columnwidth]{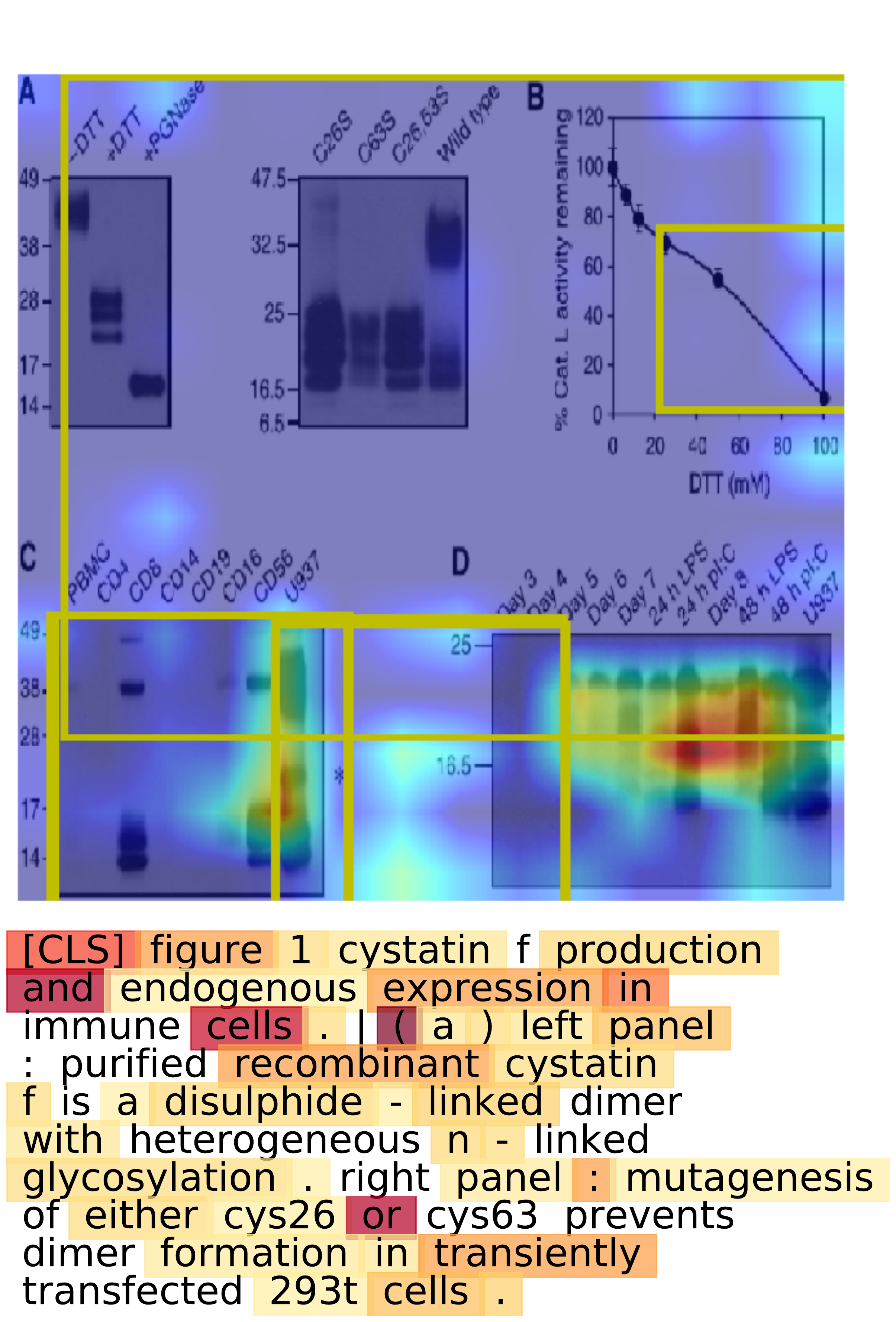} \\

Image-only \& Text-only & Multimodal & Image-only \& Text-only & Multimodal\\

\\ [-.5em]
\multicolumn{2}{c}{\textbf{(a) Interaction (Coarse)}} & \multicolumn{2}{c}{\textbf{(b) Participant (Coarse)}}

\end{tabular}

\centering
    \caption{
        \textbf{Saliency Comparisons on Int$_\text{coarse}$} (highest-lowest attention $\rightarrow$ red-blue for images and dark red-light yellow for captions): In each row: \textbf{(a)} \textit{independent} unimodal models -- ResNet-101 \& SciBERT, \textbf{(b)} multimodal model -- VL-BERT. From top to bottom the correctness of predictions between (unimodal, multimodal) is: (
        \greenchecky, \greenchecky), (\redcross, \greenchecky), and (\greenchecky, \redcross) for \textbf{coarse interaction (left half)} type, and the predictions for both unimodal models and multimodal models are \textbf{all correct} for the \textbf{coarse participant (right half)} label type. For \textbf{coarse interaction} label type, similar observations can be seen in~\figref{fig:attn_viz}, that unimodal models tend to have more dispersed attended regions, while multimodal models have more focused and finer-grained salience maps. We again show top confident ROIs in the failure case for multimodal model in the third row. Some of the ROIs are nearby or co-located with the highest attended regions, which can hypothetically cause the mis-focus and results in incorrect predictions. For \textbf{coarse participant} label type, the label space is smaller, and hence it may be easier for unimodal models to capture certain patterns to make correct predictions.
    }
    \label{fig:appendix_attn_viz}
\end{figure*}

As an extension to~\figref{fig:attn_viz} in the main paper, we provide additional visualization examples for a more in-depth look into what models have learned in~\figref{fig:appendix_attn_viz}. Similar trends are observed in both~\figref{fig:appendix_attn_viz} and~\figref{fig:attn_viz}, and we also provide the proposed ROIs from the object detection module in the third row where the multimodal model fails in the coarse \textit{interaction} label type.

Additionally, in order to further quantify how models attend differently with different modalities of inputs, we examine 20 randomly sampled data instances from the test set. We find that among which there were 13 times (65\%) that the multimodal models attended on the correct sub-figures, while image-only models only had 9 times (45\%). The difference between them is roughly the same as those results shown in~\tbref{tab:mod-comp} for the coarse-grained label types.

\subsection{Top-30 attended tokens}
\label{a-ssec:radars}

For a more in-depth understanding of what models learn from different modalities, we are interested in an overview of distributional shifts of attended words across unimodal language models and the language streams in the multimodal models. For each caption in the test set, we obtain the top-20 attended tokens by applying~\textit{SmoothGrad}~\cite{smilkov2017smoothgrad}, and then aggregate the results of every captions in the test set as an overall top-attended-token-count histogram. The results for the two types of coarse-grained labels are visualized in~\figref{fig:model_topk_radar_plots_coarse}. One can see that multimodal models (blue regions) tend to focus on more label-type-specific words, such as \textit{interaction} in~\figref{fig:model_topk_radar_plots_coarse_2}. 

Denote the frequency of a token $t_i$ in a dataset as $f_{t_i, set}$, the average relative changes of the top-$N$ attended tokens transitioning from \textit{participant} to \textit{interaction} label types is computed by: $\frac{1}{N}\sum_i^{N}[|f_{t_i, participant} - f_{t_i, interaction}| / f_{t_i, participant}]$.
Following this computation, the general trend of unimodal (language) models do not change much (the two red regions) by a 21.8$\%$ of relative changes, while it is shown that multimodal models are more susceptible to the salient decisive tokens that they exhibit relative changes by a larger margin of 33.4$\%$.

Similarly, we also visualize the top-30 attended tokens for the two finer grained labels, as shown in~\figref{fig:model_topk_radar_plots_coarse_fine}.
We find that, similar to the coarse labels, unimodal language models share similar distributional trends across the \textit{participant} and \textit{interaction} types, while multimodal models are more susceptible to label types.
The computed quantitative relative changes between the two finer-grained sets, are 30.2$\%$ and 28.1$\%$ for the multimodal models and unimodal models respectively.
For the fine-grained sets, the two relative changes are not differ by a large margin, which we hypothesize that the fine-grained sets are substantially harder for both multimodal and unimodal models to perform well.

\begin{figure}[ht!]
\begin{subfigure}{\columnwidth}
\centering
\resizebox{1.0\columnwidth}{!}{%
\begin{tikzpicture}[thick,scale=1, every node/.style={scale=2.2}]
    \coordinate (origin) at (0, 0);
    \foreach[count=\i] \counts/\dim/\color/\rwhere in {
100/protein/black/26,
76/immunoprecip/black/26,
51/antibody/black/26,
47/interaction/black/26,
42/binding/black/26,
40/western/black/26,
31/complex/black/26,
37/\shortstack{immun-\\oblot}/black/26.5,
32/extract/black/26,
29/\shortstack{immunob-\\lotting}/black/26,
27/assay/black/26,
25/lysates/black/26,
20/incubated/black/26,
17/transfected/black/26,
21/domain/black/26,
16/purified/black/26,
21/indicated/black/26,
16/yeast/black/26,
20/analysis/black/26,
18/itation/black/26,
18/hybrid/black/26,
12/mutant/black/26,
17/vitro/black/26,
7/syntax/black/26,
16/interacts/black/26,
16/using/black/26,
10/tagged/black/26,
12/strain/black/26,
10/peptide/black/26,
15/phosphorylation/black/26}
    {
        \coordinate (\i) at (\i * 360 / 30: \counts / 5);
        \node[text=\color] (title) at (\i * 360 / 30: \rwhere) {\Huge\dim};
        \draw[opacity=.5] (origin) -- (title);
    }
    \draw[fill=gray, opacity=0.03] (0, 0) circle [radius=21];
    \draw (0, 0)[opacity=.3, color=black] circle [radius=20];
    \node[opacity=.8] (title) at (5.4: 20) {\huge100};
    \draw (0, 0)[opacity=.3, color=black] circle [radius=15];
    \node[opacity=.8] (title) at (6.5: 15) {\huge75};
    \draw (0, 0)[opacity=.3, color=black] circle [radius=10];
    \node[opacity=.8] (title) at (9.8: 10) {\huge50};
    \draw (0, 0)[opacity=.3, color=black] circle [radius=5];
    \node (title) at (16: 5) {\huge25};

    \draw [fill=blue!20, opacity=.4] (1) \foreach \i in {2,...,30}{-- (\i)} --cycle;
    
    \foreach[count=\i] \counts/\dim/\color/\rwhere in {
62/protein/black/26,
61/immunoprecip/black/26,
33/antibody/black/26,
38/interaction/black/26,
35/binding/black/26,
19/western/black/26,
37/complex/black/26,
22/\shortstack{immun-\\oblot}/black/26.5,
20/extract/black/26.5,
15/\shortstack{immunob-\\lotting}/black/26,
26/assay/black/26,
13/lysates/black/26,
23/incubated/black/26,
22/transfected/black/26,
22/domain/black/26,
21/purified/black/26,
9/indicated/black/26,
20/yeast/black/26,
11/analysis/black/26,
9/itation/black/26,
13/hybrid/black/26,
17/mutant/black/26,
15/vitro/black/26,
16/syntax/black/26,
16/interacts/black/26,
5/using/black/26,
15/tagged/black/26,
15/strain/black/26,
15/peptide/black/26,
12/phosphorylation/black/26}
    {
        \coordinate (\i) at (\i * 360 / 30: \counts / 5);
    }

    \draw [fill=red!20, opacity=.4] (1) \foreach \i in {2,...,30}{-- (\i)} --cycle;
    
\end{tikzpicture}%
}
\caption{Participant Identification (Coarse)}
\label{fig:model_topk_radar_plots_coarse_1}
\end{subfigure}

\quad
\\

\begin{subfigure}{\columnwidth}
\centering
\resizebox{1.0\columnwidth}{!}{%
\begin{tikzpicture}[thick,scale=1, every node/.style={scale=2.2}]
    \coordinate (origin) at (0, 0);
    \foreach[count=\i] \counts/\dim/\color/\rwhere in {
100/immunoprecip/black/26,
49/protein/black/26,
42/antibody/black/26,
57/interaction/black/26,
51/binding/black/26,
38/\shortstack{incu-\\bated}/black/26,
20/\shortstack{trans-\\fected}/black/26,
37/\shortstack{comp-\\lex}/black/26,
35/extract/black/26,
33/itated/black/26,
32/assay/black/26,
31/itation/black/26,
30/lysates/black/26,
29/yeast/black/26,
24/immunoblot/black/26,
25/bound/black/26,
15/itates/black/26,
20/domain/black/26,
20/hybrid/black/26,
15/immunoblotting/black/26,
12/tagged/black/26,
13/mutant/black/26,
17/western/black/26,
17/vitro/black/26,
17/analysis/black/26,
16/fraction/black/26,
11/purified/black/26,
11/activity/black/26,
14/stained/black/26,
12/phosphorylation/black/26}
    {
        \coordinate (\i) at (\i * 360 / 30: \counts / 5);
        \node[text=\color] (title) at (\i * 360 / 30: \rwhere) {\Huge\dim};
        \draw[opacity=.5] (origin) -- (title);
    }
    \draw[fill=gray, opacity=0.03] (0, 0) circle [radius=21];
    \draw (0, 0)[opacity=.3, color=black] circle [radius=20];
    \node[opacity=.8] (title) at (5.4: 20) {\huge100};
    \draw (0, 0)[opacity=.3, color=black] circle [radius=15];
    \node[opacity=.8] (title) at (6.5: 15) {\huge75};
    \draw (0, 0)[opacity=.3, color=black] circle [radius=10];
    \node[opacity=.8] (title) at (9.8: 10) {\huge50};
    \draw (0, 0)[opacity=.3, color=black] circle [radius=5];
    \node (title) at (16: 5) {\huge25};

    \draw [fill=blue!20, opacity=.4] (1) \foreach \i in {2,...,30}{-- (\i)} --cycle;
    
    \foreach[count=\i] \counts/\dim/\color/\rwhere in {
92/immunoprecip/black/26,
69/protein/black/26,
59/antibody/black/26,
34/interaction/black/26,
39/binding/black/26,
21/incubated/black/26,
37/transfected/black/26,
31/complex/black/26,
21/extract/black/26,
19/itated/black/26,
25/assay/black/26,
19/itation/black/26,
16/lysates/black/26,
24/yeast/black/26,
28/immunoblot/black/26,
15/bound/black/26,
20/itates/black/26,
15/domain/black/26,
14/hybrid/black/26,
19/immunoblotting/black/26,
18/tagged/black/26,
17/mutant/black/26,
14/western/black/26,
11/vitro/black/26,
13/analysis/black/26,
13/fraction/black/26,
15/purified/black/26,
14/activity/black/26,
13/stained/black/26,
13/phosphorylation/black/26}
    {
        \coordinate (\i) at (\i * 360 / 30: \counts / 5);
    }

    \draw [fill=red!20, opacity=.4] (1) \foreach \i in {2,...,30}{-- (\i)} --cycle;
    
\end{tikzpicture}%
}
\caption{Interaction Detection (Coarse)}
\label{fig:model_topk_radar_plots_coarse_2}
\end{subfigure}

\caption{\textbf{Top-30 attended tokens across whole test set on \textit{coarse} label types}
(blue regions of VL-BERT, and red regions of SciBERT.):
For each caption, we compute the top-20 attended tokens using \textit{SmoothGrad}, and then combine the top-attended token counts from all the captions in the test set for an overall most attended (top-30) tokens, with lemmatization applied. The token counts are normalized w.r.t the maximum values across the two models.
}
\label{fig:model_topk_radar_plots_coarse}
\end{figure}
\begin{figure}[ht]
\begin{subfigure}{\columnwidth}
\centering
\resizebox{1.0\columnwidth}{!}{%
\begin{tikzpicture}[thick,scale=1, every node/.style={scale=2.2}]
    \coordinate (origin) at (0, 0);
    \foreach[count=\i] \counts/\dim/\color/\rwhere in {
69/immunoprecip/black/26,
49/protein/black/26,
17/transfected/black/26,
35/antibody/black/26,
19/binding/black/26,
33/interaction/black/26,
31/\shortstack{ext-\\ract}/black/26.5,
23/western/black/26,
29/immunoblot/black/26,
24/immunoblotting/black/26,
23/lysates/black/26,
23/complex/black/26,
22/yeast/black/26,
21/assay/black/26,
18/tagged/black/26,
11/incubated/black/26,
15/analysis/black/26,
12/domain/black/26,
16/hybrid/black/26,
12/mutant/black/26,
9/stained/black/26,
9/plasmid/black/26,
14/\shortstack{int-\\eracts}/black/26,
7/analysed/black/26,
13/itation/black/26,
13/blotting/black/26,
11/phosphorylation/black/26,
12/indicated/black/26,
7/panel/black/26,
11/vitro/black/26}
    {
        \coordinate (\i) at (\i * 360 / 30: \counts / 5);
        \node[text=\color] (title) at (\i * 360 / 30: \rwhere) {\Huge\dim};
        \draw[opacity=.5] (origin) -- (title);
    }
    \draw[fill=gray, opacity=0.03] (0, 0) circle [radius=21];
    \draw (0, 0)[opacity=.3, color=black] circle [radius=20];
    \node[opacity=.8] (title) at (5.4: 20) {\huge100};
    \draw (0, 0)[opacity=.3, color=black] circle [radius=15];
    \node[opacity=.8] (title) at (6.5: 15) {\huge75};
    \draw (0, 0)[opacity=.3, color=black] circle [radius=10];
    \node[opacity=.8] (title) at (9.8: 10) {\huge50};
    \draw (0, 0)[opacity=.3, color=black] circle [radius=5];
    \node (title) at (16: 5) {\huge25};

    \draw [fill=blue!20, opacity=.5] (1) \foreach \i in {2,...,30}{-- (\i)} --cycle;
    
    \foreach[count=\i] \counts/\dim/\color/\rwhere in {
100/immunoprecip/black/26,
49/protein/black/26,
41/transfected/black/26,
32/antibody/black/26,
33/binding/black/26,
32/interaction/black/26,
8/extract/black/26,
29/western/black/26,
29/immunoblot/black/26,
23/immunoblotting/black/26,
18/lysates/black/26,
22/complex/black/26,
17/yeast/black/26,
18/assay/black/26,
20/tagged/black/26,
20/incubated/black/26,
17/analysis/black/26,
16/domain/black/26,
12/hybrid/black/26,
15/mutant/black/26,
14/stained/black/26,
14/plasmid/black/26,
9/interacts/black/26,
13/analysed/black/26,
10/itation/black/26,
10/blotting/black/26,
12/phosphorylation/black/26,
6/indicated/black/26,
11/panel/black/26,
10/vitro/black/26}
    {
        \coordinate (\i) at (\i * 360 / 30: \counts / 5);
    }

    \draw [fill=red!20, opacity=.3] (1) \foreach \i in {2,...,30}{-- (\i)} --cycle;
    
\end{tikzpicture}%
}
\caption{Participant Identification (Fine)}
\label{fig:model_topk_radar_plots_coarse_fine_1}
\end{subfigure}

\quad
\\

\begin{subfigure}{\columnwidth}
\centering
\resizebox{1.0\columnwidth}{!}{%
\begin{tikzpicture}[thick,scale=1, every node/.style={scale=2.2}]
    \coordinate (origin) at (0, 0);
    \foreach[count=\i] \counts/\dim/\color/\rwhere in {
89/immunoprecip/black/26,
52/antibody/black/26,
48/protein/black/26,
43/interaction/black/26,
41/binding/black/26,
39/complex/black/26,
26/\shortstack{trans-\\fected}/black/26,
34/\shortstack{incu-\\bated}/black/26,
33/itated/black/26,
33/assay/black/26,
30/yeast/black/26,
23/lysates/black/26,
22/extract/black/26,
24/bound/black/26,
16/immunoblotting/black/26,
21/tagged/black/26,
23/itation/black/26,
21/hybrid/black/26,
17/immunoblot/black/26,
19/using/black/26,
17/domain/black/26,
13/mutant/black/26,
17/analysis/black/26,
17/purified/black/26,
16/vitro/black/26,
14/phosphorylation/black/26,
14/itates/black/26,
15/immobilized/black/26,
10/stained/black/26,
13/endogenous/black/26}
    {
        \coordinate (\i) at (\i * 360 / 30: \counts / 5);
        \node[text=\color] (title) at (\i * 360 / 30: \rwhere) {\Huge\dim};
        \draw[opacity=.5] (origin) -- (title);
    }
    \draw[fill=gray, opacity=0.03] (0, 0) circle [radius=21];
    \draw (0, 0)[opacity=.3, color=black] circle [radius=20];
    \node[opacity=.8] (title) at (5.4: 20) {\huge100};
    \draw (0, 0)[opacity=.3, color=black] circle [radius=15];
    \node[opacity=.8] (title) at (6.5: 15) {\huge75};
    \draw (0, 0)[opacity=.3, color=black] circle [radius=10];
    \node[opacity=.8] (title) at (9.8: 10) {\huge50};
    \draw (0, 0)[opacity=.3, color=black] circle [radius=5];
    \node (title) at (16: 5) {\huge25};

    \draw [fill=blue!20, opacity=.5] (1) \foreach \i in {2,...,30}{-- (\i)} --cycle;
    
    \foreach[count=\i] \counts/\dim/\color/\rwhere in {
100/immunoprecip/black/26,
94/antibody/black/26,
76/protein/black/26,
26/interaction/black/26,
42/binding/black/26,
25/complex/black/26,
35/transfected/black/26,
30/incubated/black/26,
29/itated/black/26,
32/assay/black/26,
26/yeast/black/26,
27/lysates/black/26,
24/extract/black/26,
21/bound/black/26,
23/immunoblotting/black/26,
23/tagged/black/26,
20/itation/black/26,
19/hybrid/black/26,
20/immunoblot/black/26,
10/using/black/26,
18/domain/black/26,
17/mutant/black/26,
14/analysis/black/26,
15/purified/black/26,
16/vitro/black/26,
16/phosphorylation/black/26,
15/itates/black/26,
10/immobilized/black/26,
13/stained/black/26,
9/endogenous/black/26}
    {
        \coordinate (\i) at (\i * 360 / 30: \counts / 5);
    }

    \draw [fill=red!20, opacity=.3] (1) \foreach \i in {2,...,30}{-- (\i)} --cycle;
    
\end{tikzpicture}%
}
\caption{Interaction Detection (Fine)}
\label{tab:model_topk_radar_plots_coarse_fine_2}
\end{subfigure}

\caption{\textbf{Top-30 attended tokens across whole test set on \textit{fine} label types}
(blue regions of VL-BERT, and red regions of SciBERT). Similar to~\figref{fig:model_topk_radar_plots_coarse} but on the fine-grained label types.
}
\label{fig:model_topk_radar_plots_coarse_fine}
\end{figure}

\section{Releases}
\label{a-sec: impl}
The full dataset along with its documentation will be released in a timely manner.
We will release the cleaned code repository which encompasses the majority of the codes used to generate the experimental results in this paper.
The repository will also include codes for the VL-BERT model, adapted from the original publicly available repository, for exemplifying how we adapt the existing models to our dataset and setups.
We will also release our \textit{distantly supervised} data collection tool (all components of the three main steps in our data collection pipeline), with well documented guidelines of usages. We hope by sharing these tools, more interest could be gained into developing better multimodal models especially for traditionally low resource areas such as biomedical domains.

\end{document}